\documentclass{article}

\usepackage{microtype}
\usepackage{graphicx}
\usepackage{subfigure}
\usepackage{booktabs} % for professional tables

\usepackage{hyperref}

\usepackage[accepted]{icml2023}

\usepackage{amsmath}
\usepackage{amssymb}
\usepackage{mathtools}
\usepackage{amsthm}

\usepackage[capitalize,noabbrev]{cleveref}

\theoremstyle{plain}

\theoremstyle{definition}

\theoremstyle{remark}

\usepackage[textsize=tiny]{todonotes}

\icmltitlerunning{Beam Tree Recursive Cells}

\begin{document}

\twocolumn[
\icmltitle{Beam Tree Recursive Cells}

% It is OKAY to include author information, even for blind
% submissions: the style file will automatically remove it for you
% unless you've provided the [accepted] option to the icml2023
% package.

% List of affiliations: The first argument should be a (short)
% identifier you will use later to specify author affiliations
% Academic affiliations should list Department, University, City, Region, Country
% Industry affiliations should list Company, City, Region, Country

% You can specify symbols, otherwise they are numbered in order.
% Ideally, you should not use this facility. Affiliations will be numbered
% in order of appearance and this is the preferred way.
\icmlsetsymbol{equal}{*}

\begin{icmlauthorlist}
\icmlauthor{Jishnu Ray Chowdhury}{yyy}
\icmlauthor{Cornelia Caragea}{yyy}
\end{icmlauthorlist}

\icmlaffiliation{yyy}{Computer Science, University of Illinois Chicago}

\icmlcorrespondingauthor{Jishnu Ray Chowdhury}{jraych2@uic.edu}
\icmlcorrespondingauthor{Cornelia Caragea}{cornelia@uic.edu}

% You may provide any keywords that you
% find helpful for describing your paper; these are used to populate
% the "keywords" metadata in the PDF but will not be shown in the document
\icmlkeywords{Machine Learning, ICML}

\vskip 0.3in
]

% this must go after the closing bracket ] following \twocolumn[ ...

% This command actually creates the footnote in the first column
% listing the affiliations and the copyright notice.
% The command takes one argument, which is text to display at the start of the footnote.
% The \icmlEqualContribution command is standard text for equal contribution.
% Remove it (just {}) if you do not need this facility.

\printAffiliationsAndNotice{}  % leave blank if no need to mention equal contribution
%\printAffiliationsAndNotice{\icmlEqualContribution} % otherwise use the standard text.

\begin{abstract}
We propose Beam Tree Recursive Cell (BT-Cell) - a backpropagation-friendly framework to extend Recursive Neural Networks (RvNNs) with beam search for latent structure induction. We further extend this framework by proposing a relaxation of the hard top-$k$ operators in beam search for better propagation of gradient signals. We evaluate our proposed models in different out-of-distribution splits in both synthetic and realistic data. Our experiments show that BT-Cell achieves near-perfect performance on several challenging structure-sensitive synthetic tasks like ListOps and logical inference while maintaining comparable performance in realistic data against other RvNN-based models.  Additionally, we identify a previously unknown failure case for neural models in generalization to unseen number of arguments in ListOps. The code is available at: \url{https://github.com/JRC1995/BeamTreeRecursiveCells}. 
\end{abstract}

\section{Introduction}
In the space of sequence encoders, Recursive Neural Networks (RvNNs) can be said to lie somewhere in-between Recurrent Neural Networks (RNNs) and Transformers in terms of flexibility. While vanilla Transformers show phenomenal performance and  scalability on a variety of tasks, they can often struggle in length generalization and systematicity in syntax-sensitive tasks \citep{tran-2018-importance, shen-2019-ordered, Lakretz2021CausalTP, csordas-2022-ndr}. RvNN-based models, on the other hand, can often excel on some of the latter kind of tasks \citep{shen-2019-ordered, chowdhury-2021-modeling, liu-etal-2021-learning-algebraic, bogin-etal-2021-latent} making them worthy of further study although they may suffer from limited scalability in their current formulations. 

Given an input text, RvNNs \citep{pollack-1990-recursive, goller1996learning, Socher10learningcontinuous} are designed to build up the representation of the whole text by recursively building up the representations of their constituents starting from the most elementary representations (tokens) in a bottom-up fashion. As such, RvNNs can model the hierarchical part-whole structures underlying texts. However, originally RvNNs required access to pre-defined hierarchical constituency-tree structures. Several works \citep{choi-2018-learning, peng-etal-2018-backpropagating, havrylov-2019-cooperative, maillard_clark_yogatama_2019, chowdhury-2021-modeling} introduced latent-tree RvNNs that sought to move beyond this limitation by making RvNNs able to learn to automatically determine the structure of composition from any arbitrary downstream task objective. 

Among these approaches, Gumbel-Tree models \citep{choi-2018-learning} are particularly attractive for their simplicity. However, they not only suffer from biased gradients due to the use of Straight-Through Estimation (STE) \cite{bengio-2013-estimating}, but they also perform poorly on synthetic tasks like ListOps \citep{nangia-2018-listops,williams-etal-2018-latent} that were specifically designed to diagnose the capacity of neural models for automatically inducing underlying hierarchical structures. To tackle these issues, we propose the Beam Tree Cell (BT-Cell) framework that incorporates beam-search on RvNNs replacing the STE Gumbel Softmax \cite{jang-2017-categorical, maddison2017the} in Gumbel-Tree models. Instead of greedily selecting the highest scored sub-tree representations like Gumbel-Tree models, BT-Cell chooses and maintains top-$k$ highest scored sub-tree representations.  We show that BT-Cell outperforms Gumbel-Tree models in challenging structure sensitive tasks by several folds. For example, in ListOps, when testing for samples of length $900$-$1000$, BT-Cell increases the performance of a comparable Gumbel-Tree model from $37.9\%$ to $86.7\%$ (see Table \ref{table:listops}). We further extend BT-Cell by replacing its non-differentiable top-$k$ operators with a novel operator called OneSoft Top-$k$. Our proposed operator, combined with BT-Cell, achieves a new state-of-the-art in length generalization and depth-generalization in structure-sensitive synthetic tasks like ListOps and performs comparably in realistic data against other strong models. 

A few recently proposed latent-tree models simulating RvNNs including Tree-LSTM-RL \citep{havrylov-2019-cooperative}, Ordered Memory (OM) \citep{shen-2019-ordered} and Continuous RvNNs (CRvNNs) \citep{chowdhury-2021-modeling} are also strong contenders to BT-Cell on synthetic data. However, unlike BT-Cell, Tree-LSTM-RL relies on reinforcement learning and several auxiliary techniques to stabilize training. Moreover, compared to OM and CRvNN, one distinct advantage of BT-Cell is that it does not just provide the final sequence encoding (representing the whole input text) but also the intermediate constituent representations at different levels of the hierarchy (representations of all nodes of the underlying induced trees). Such tree-structured node representations can be useful as inputs to further downstream modules like a Transformer \citep{vaswani2017} or Graph Neural Network \citep{scarselli2000graph} in a full end-to-end setting.\footnote{There are several works that have used intermediate span representations for better compositional generalization in generalization tasks \citep{liu-2020-compo, herzig-berant-2021-span, bogin-etal-2021-latent, liu-etal-2021-learning-algebraic,mao2021grammarbased}. We keep it as a future task to explore whether the span representations returned by BT-Cell can be used in relevant ways.} While CYK-based RvNNs \citep{maillard_clark_yogatama_2019} are also promising and similarly can provide multiple span representations they tend to be much more expensive than BT-Cell (see $\S$\ref{efficiency}).

As a further contribution, we also identify a previously unknown failure case for even the best performing neural models when it comes to argument generalization in ListOps \citep{nangia-2018-listops}---opening up a new challenge for future research.

\section{Preliminaries}
\textbf{Problem Formulation:} Similar to \citet{choi-2018-learning}, throughout this paper, we explore the use of RvNNs as a sentence encoder. Formally, given a sequence of token embeddings $\mathcal{X} = (e_1, e_2, \dots, e_n)$ (where $\mathcal{X} \in {\rm I\!R}^{n \times d_e}$ and  $e_i \in {\rm I\!R}^{d_e}$; $d_e$ being the embedding size), the task of a sentence encoding function $\mathcal{E}:{\rm I\!R}^{n \times d_e} \rightarrow {\rm I\!R}^{d_h}$ is to encode the whole sequence of vectors into a single vector $o = \mathcal{E}(\mathcal{X})$ (where $o \in {\rm I\!R}^{d_h}$ and $d_h$ is the size of the encoded vector). We can use a sentence encoder for sentence-pair comparison tasks like logical inference or for text classification.

\subsection{RNNs and RvNNs}
\label{rnn}
A core component of both RNNs and RvNNs is a recursive cell $R$. In our context, $R$ takes as arguments two vectors ($a_1  \in {\rm I\!R}^{d_{a_1}}$ and $a_2  \in {\rm I\!R}^{d_{a_2}}$)  and returns a single vector $v = R(a_1,a_2)$. $R: {\rm I\!R}^{d_{a_1}} \times {\rm I\!R}^{d_{a_2}}  \rightarrow {\rm I\!R}^{d_v}$. In our settings, we generally set $d_{a_1} = d_{a_2} = d_{v} = d_h$. Given a sequence $\mathcal{X}$, both RNNs and RvNNs sequentially process it through a recursive application of the cell function. For a concrete example, consider a sequence of token embeddings such as $(2+4 \times 4+3)$ (assume the symbols $2$, $4$, $+$ etc. represent the corresponding embedding vectors $\in {\rm I\!R}^{d_h}$). Given any such sequence, RNNs can only follow a fixed left-to-right order of composition. For the particular aforementioned sequence, an RNN-like application of the cell function can be expressed as: 
\begin{equation}
o = R(R(R(R(R(R(R(h0,2),+),4),\times),4),+),3)
\end{equation}
Here, $h0$ is the initial hidden state. In contrast to RNNs, RvNNs can compose the sequence in more flexible orders. For example, one way (among many) that RvNNs could apply the cell function is as follows:
\begin{equation}
o = R(R(R(R(2,+),R(R(4,\times),4)),+),3)
\label{rvnn}
\end{equation}
Thus, RvNNs can be considered as a generalization of RNNs where a strict left-to-right order of composition is not anymore enforced. As we can see, by this strategy of recursively reducing two vectors into a single vector, both RNNs and RvNNs can implement the sentence encoding function in the form of $\mathcal{E}$. Moreover, the form of application of cell function exhibited by RNNs and RvNNs can also be said to reflect a tree-structure. For any application of the cell function in the form $v=R(a_1, a_2)$, $v$ can be treated as the representation of the immediate parent node of child nodes $a_1$ and $a_2$ in an underlying tree.

In Eqn. \ref{rvnn}, we find that RvNNs can align the order of composition to PEMDAS whereas RNNs cannot. Nevertheless, RNNs can still learn to simulate RvNNs by modeling tree-structures implicitly in their hidden state dimensions \citep{bowman-2015-tree}. For example, RNNs can learn to hold off the information related to ``$2 +$" until ``$4 \times 4$" is processed. Their abilities to handle tree-structures is analogous to how we can use pushdown automation in a recurrent manner through an infinite stack to detect tree-structured grammar. Still, RNNs can struggle to effectively learn to appropriately organize information in practice for large sequences. Special inductive biases can be incorporated to enhance their abilities to handle their internal memory structures \citep{shen-2018-ordered, shen-2019-ordered}. However, even then, memories remain bounded in practice and there is a limit to what depth of nested structures they can model. 

More direct approaches to RvNNs, in contrast, can alleviate the above problems and mitigate the need of sophisticated memory operations to arrange information corresponding to a tree-structure because they can directly compose according to the underlying structure (Eqn. \ref{rvnn}). However, in the case of RvNNs, we have the problem of first determining the underlying structure to even start the composition. One approach to handle the issue can be to train a separate parser to induce a tree structure from sequences using gold tree parses. Then we can use the trained parser in RvNNs. However, this is not ideal. Not all tasks or languages would come with gold trees for training a parser and a parser trained in one domain may not translate well to another. A potentially better approach is to jointly learn both the cell function and structure induction from a downstream objective \cite{choi-2018-learning}. We focus on this latter approach. Below we discuss one framework (easy-first parsing and Gumbel-Tree models) for this approach. 

\subsection{Easy-First Parsing and Gumbel-Tree Models}
\label{easy-first}
Here we describe an adaptation \citep{choi-2018-learning} of easy-first parsing \citep{goldberg-2010-efficient} for RvNN-based sentence-encoding. The algorithm relies on a scorer function $score: {\rm I\!R}^{d_h} \rightarrow {\rm I\!R}^1$ that scores parsing decisions. Particularly, if we have $v = R(a_1,a_2)$, then $score(v)$ represents the plausibility of $a_1$ and $a_2$ belonging to the same immediate parent constituent. Similar to \cite{choi-2018-learning}, we keep the scorer as a simple linear transformation: $score(v) = vW_v$ (where $W_v \in {\rm I\!R}^{d_h \times 1}$ and $v \in {\rm I\!R}^{d_h}$).  

\textbf{Recursive Loop:} In this algorithm, at every iteration in a recursive loop, given a sequence of hidden states $(h_1, h_2,\dots,h_n)$ we consider all possible immediate candidate parent compositions taking the current states as children: $(R(h_1, h_2), R(h_2, h_3),\dots,R(h_{n-1},h_n))$.\footnote{We focus only on the class of binary projective tree structures. Thus all the candidates are compositions of two contiguous elements.} We then score each of the candidates with the score function and greedily select the highest scoring candidate (i.e., we commit to the ``easiest" decision first). For the sake of illustration, assume $score(R(h_i,h_{i+1})) \geq score(R(h_j,h_{j+1})) \; \forall j \in \{1,2,\dots, n\}$. Thus, following the algorithm, the parent candidate $R(h_i,h_{i+1})$ will be chosen. The parent representation $R(h_i,h_{i+1})$ would then replace its immediate children $h_i$ and $h_{i+1}$. Thus, the resulting sequence will become: $(h_1,\dots,h_{i-1},R(h_i,h_{i+1}),h_{i+2},\dots,h_n)$. Like this, the sequence will be iteratively reduced to a single element representing the final sentence encoding. The full algorithm is presented in the Appendix (see Algorithm \ref{alg:easy-first}).    

One issue here is to decide how to choose the highest scoring candidate. One way to do this is to simply use an argmax operator but it will not be differentiable. Gumbel-Tree models \citep{choi-2018-learning} address this by using Straight Through Estimation (STE) \citep{bengio-2013-estimating} with Gumbel Softmax \citep{jang-2017-categorical, maddison2017the} instead of argmax. However, STE is known to cause high bias in gradient estimation. Moreover, as it was previously discovered \citep{nangia-2018-listops}, and as we independently verify, STE Gumbel-based strategies perform poorly when tested in structure-sensitive tasks. Instead, to overcome these issues, we propose an alternative of extending argmax with a top-$k$ operator under a beam search strategy.  

\section{Beam Tree Cell}
\label{bt-cell}
\textbf{Motivation:} Gumbel-Tree models, as described, are relatively fast and simple but they are fundamentally based on a greedy algorithm for a task where the greedy solution is not guaranteed to be optimal. On the other hand, adaptation of dynamic programming-based CYK-models \citep{maillard_clark_yogatama_2019} leads to high computational complexity (see $\S$\ref{efficiency}). A ``middle way" between the two extremes is then to simply extend Gumbel-Tree models with beam-search to make them less greedy while still being less costly than CYK-parsers. Moreover, using beam-search also provides additional opportunity to recover from local errors whereas a greedy single-path approach (like Gumbel-Tree models) will be stuck with any errors made. All these factors motivate the framework of Beam Tree Cells (BT-Cell). 

\textbf{Implementation:} The beam search extension to Gumbel-Tree models is straight-forward and similar to standard beam search. The method is described more precisely in Appendix \ref{btcell-algo} and Algorithm \ref{alg:bt-cell}. In summary, in BT-Cell, given a beam size $k$, we maintain a maximum of $k$ hypotheses (or beams) at each recursion. In any given iteration, each beam constitutes a sequence of hidden states representing a particular path of composition and an associated score for that beam based on the addition of log-softmaxed outputs of the $score$ function (as defined in $\S$\ref{easy-first}) over each chosen composition for that sequence. At the end of the recursion, we will have $k$ sentence encodings ($(o_1,o_2,\dots,o_k)$ where $o_i \in {\rm I\!R}^{d_h}$) and their corresponding scores ($(s_1,s_2,\dots,s_k)$ where $s_i \in {\rm I\!R}^{1}$). The final sequence encoding can be then represented as: 
% \begin{equation}
$\sum_{i=1}^k \left(\frac{exp(s_i) \cdot o_i}{\sum_{i=1}^k exp(s_i)}\right)$.
% \end{equation}
This aims at computing the expectation over the $k$ sequence encodings. 

\begin{figure*}[t]
    \centering
    \includegraphics[scale=0.2]{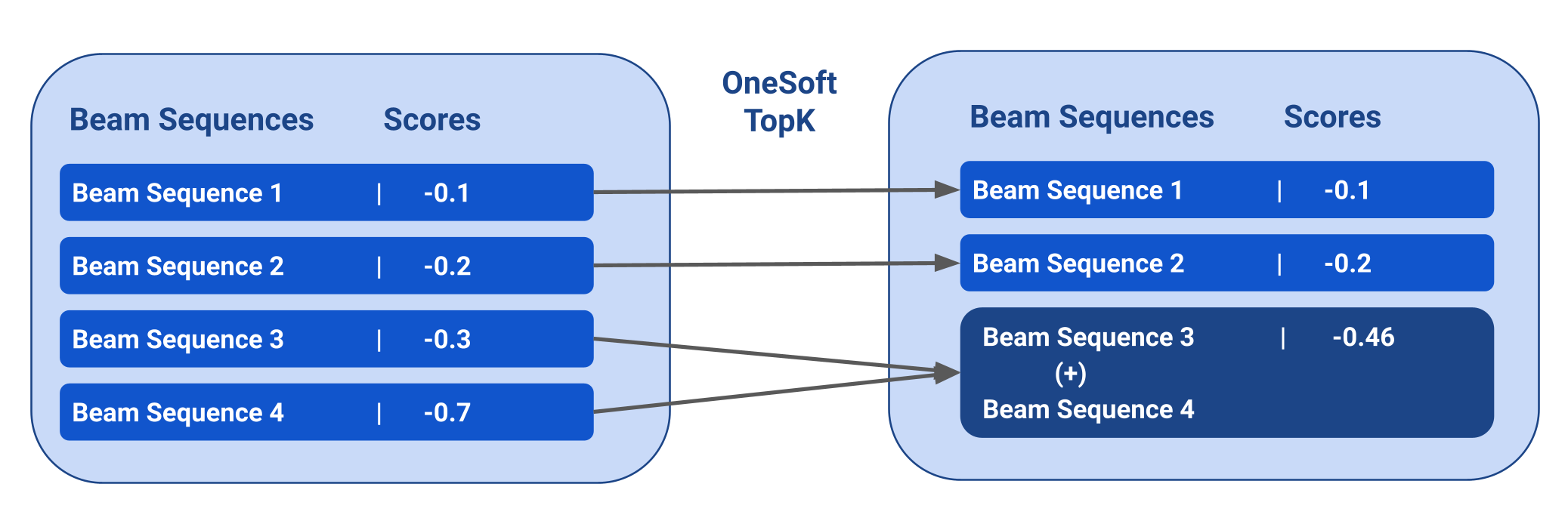}
    \caption{Visualization of OneSoft Top-$k$ selection from $m=4$ beams to top $k=3$ beams. (+) represents interpolation. 
    }
    \label{fig:softpath}
\end{figure*}

\subsection{Top $k$ Variants}
\label{topks}
As in standard beam search, BT-Cell requires two top-$k$ operators. The first top-$k$ replaces the straight-through Gumbel Softmax (simulating top-1) in Gumbel-Tree models. However, selecting and maintaining $k$ possible choices for every beam in every iteration leads to an exponential increase in the number of total beams. Thus, a second top-$k$ operator is used for pruning the beams to maintain only a maximum of $k$ beams at the end of each iteration. Here, we focus on variations of the second top-$k$ operator that is involved in truncating beams.

\textbf{Plain Top-$k$:} The simplest variant is just the vanilla top-$k$ operator. However, the vanilla top-$k$ operator is discrete and non-differentiable preventing gradient propagation to non-selected paths.\footnote{Strictly speaking, in practice, we generally use stochastic top-$k$ \cite{kool2019stochastic} during training but in our preliminary experiments we did not find this choice to bear much weight.} Despite that, this can still work for the following reasons: (1) gradients can still pass through the final top $k$ beams and scores. The scorer function can thus learn to increase the scores of better beams and lower the scores of the worse ones among the final $k$ beams; (2) a rich enough cell function can be robust to local errors in the structure and learn to adjust for it by organizing information better in its hidden states. We believe that as a combination of these two factors, plain BT-Cell even with non-differentiable top-$k$ operators can learn to perform well for structure-sensitive tasks (as we will empirically observe).

\textbf{OneSoft Top-$k$:} While non-differentiable top-$k$ operators can work, they still can be a bottleneck because gradient signals will be received only for $k$ beams in a space of exponential possibilities. To address this issue, we consider if we can make the truncation or deletion of beams ``softer". To that end, we develop a new Top-$k$ operator that we call OneSoft Top-$k$. We motivate and describe it below.

As a concrete case, assume we have $m$ beams (sequences and their corresponding scores). The target for a top-$k$ operator is to keep only the top scoring $k$ beams (where $k \leq m$).  Ideally we want to keep the beam representations ``sharp" and avoid washed out representations owing to interpolation (weighted vector averaging) of distinct paths \citep{drozdov-etal-2020-unsupervised}. This can be achieved by plain top-$k$. However, it prevents propagation of gradient signals through the bottom $m-k$ beams. Another line of approach is to create a soft permutation matrix $P \in {\rm I\!R}^{m \times m}$ through a differentiable sorting algorithm such that $P_{ij}$ represents the probability of the $i^{th}$ beam being the $j^{th}$ highest scoring beam. $P$ can then be used to softly select the top $k$ beams. However, running differentiable sorting in a recursive loop can significantly increase computation overheads and can also create more ``washed out" representations leading to higher error accumulation (also see $\S$\ref{ablations} and $\S$\ref{efficiency}). We tackle all these challenges by instead proposing OneSoft as a simple hybrid strategy to approach top-$k$ selection. We provide a formal description of our proposed strategy below and a visualization of the process in Figure \ref{fig:softpath}. 

Assume we have $m$ beams consisting of $m$ sequences: $H = (\mathcal{H}_1, \dots, \mathcal{H}_m)$ ($\mathcal{H}_i \in {\rm I\!R}^{n \times d_h}$ where $n$ is the sequence length) and $m$ corresponding scalar scores: $S=(s_1, \dots, s_m)$. First, we simply use the plain top-$k$ operator to discretely select the top $k-1$ beams (instead of $k$). This allows us to keep the most promising beams ``sharp": 
\begin{equation}
    idx = topk(S, K=k-1),\;\;\;\;\textit{Top} = \{(\mathcal{H}_i, s_i)\text{ }|\text{ }i \in idx\}
\end{equation}
Second, for the $k^{th}$ beam we instead perform a softmax-based marginalization of the bottom $m-(k-1)$ beams. This allows us to still propagate gradients through the bottom scoring beams (unlike in the pure plain top-$k$ operator):
\begin{equation}
    \textit{B} = \{(\mathcal{H}_i, s_i)\text{ }|\text{ }(i \notin idx) \land (i \in \{1,2,\dots,m\})\}
\end{equation}
\begin{equation}
    Z = \sum_{(\mathcal{H},s) \in B} \exp(s) 
\end{equation}
\begin{equation}
    \textit{SP} = \left(\sum_{(\mathcal{H}, s) \in B} \left(\frac{\exp(s)}{Z}\cdot \mathcal{H}\right) , \sum_{(\mathcal{H}, s) \in B} \left(\frac{exp(s)}{Z} \cdot s\right)  \right)
    \label{eqn:softpath}
\end{equation}
Here $B$ represents the bottom scoring $m-(k-1)$ beams and $SP$ represents the softmax-based marginalization. Finally, we add the $SP$ to the top $k-1$ discretely selected beams to get the final set of $k$ beams: $\textit{Top} \cup \{SP\}$. Thus, we get to achieve a ``middle way" between plain top-$k$ and differentiable sorting: partially getting the benefit of sharp representations of the former through discrete top $k-1$ selection, and partially getting the benefit of gradient propagation of the latter through soft-selection of the $k^{th}$ beam. In practice, we switch to plain top-$k$ during inference. This makes tree extraction convenient during inference if needed.

\begin{table*}[t]
\small
\centering
\def\arraystretch{1.2}
\begin{tabular}{  l | l | l l l | l l |l} 
%\toprule
\hline
\textbf{Model} & \textbf{near-IID} & \multicolumn{3}{c}{\textbf{Length Gen.}} & \multicolumn{2}{|c|}{\textbf{Argument Gen.}} & \textbf{LRA}\\
(Lengths) & $\leq$ 1000 & 200-300 & 500-600 & 900-1000 & 100-1000 & 100-1000 & $2000$\\
(Arguments) & $\leq$ 5 & $\leq$ 5 & $\leq$ 5 & $\leq$ 5 & 10 & 15 & 10\\
\hline
\multicolumn{5}{l}{\textit{With gold trees}}\\
\hline
GoldTreeGRC & $99.95$ & $99.88$ & $99.85$ & $100$ & $80.5$ & $79$ & $78.1$\\
\hline
\multicolumn{5}{l}{\textit{Baselines without gold trees}}\\
\hline
RecurrentGRC & $84.05$ & $33.85$ & $20.2$ & $15.1$ & $37.35$ & $30.10$ & $20.7$\\
GumbelTreeGRC & $74.89$ & $47.6$ & $43.85$ & $37.9$ & $51.35$ & $50.5$ & $46.1$\\
CYK-GRC & $97.87$ & $93.75$ & --- & --- & $60.75$ & $42.45$ & ---\\
Ordered Memory & $\underline{99.88}$ & $\mathbf{99.55}$ & $92.7$ & $76.9$ & $\mathbf{84.15}$ & $\mathbf{75.05}$ & $\mathbf{80.1}$\\
%CRvNN$\dagger$ & $99.6_3$ & $98.51_{11}$ & $97.95_{11}$ & $96.78_{19}$ & --- & --- & ---\\
%CRvNN & $98.86$ & $95.89$ & $93.15$ & $89.4$ & $57.8$ & $24.35$ & $45.1$ \\
CRvNN & $99.82$ & $\underline{99.5}$ & $98.5$ & $\mathbf{98}$ & $65.45$ & $45.1$ & $55.38$\\
\hline
\multicolumn{5}{l}{\textit{Ours}}\\
\hline
BT-GRC & $99.39$ & $96.15$ & $92.55$ & $86.7$ & $\underline{77.1}$ & $63.7$ & $67.3$\\
BT-GRC + OneSoft & $\mathbf{99.92}$ & $\underline{99.5}$ & $\mathbf{99}$ & $\underline{97.2}$ & $76.05$ & $\underline{67.9}$ & $\underline{71.8}$\\
\hline
\end{tabular}
\caption{Accuracy on ListOps. For our models we report the median of $3$ runs except for CYK-GRC which was ran only once for its high resource demands. Our models were trained on lengths $\leq$ 100, depth $\leq$ 20, and arguments $\leq$ 5. We bold the best results and underline the second-best among models that do not use gold trees.}
\label{table:listops}
\vspace{-4mm}
\end{table*}

\section{Experiments and Results}

We present the main models below. Hyperparameters and other architectural details are in Appendix \ref{hyperparameters}. 

\textbf{1. RecurrentGRC:} RecurrentGRC is an RNN implemented with the Gated Recursive Cell (GRC) \citep{shen-2019-ordered}  as the cell function (see Appendix \ref{grc} for description of GRC). \textbf{2. GoldTreeGRC:} GoldTreeGRC is a GRC-based RvNN with gold tree structures. \textbf{3. GumbelTreeGRC:} This is the same as GumbelTreeLSTM \cite{choi-2018-learning} but with GRC instead of LSTM. \textbf{4. CYK-GRC:} This is the CYK-based model proposed by \citet{maillard_clark_yogatama_2019} but with GRC. \textbf{5. Ordered Memory (OM):} This is a memory-augmented RNN simulating certain classes of RvNN functions as proposed by \citet{shen-2019-ordered}. OM also uses GRC. \textbf{6. CRvNN:} CRvNN is a variant of RvNN with a continuous relaxation over its structural operations as proposed by \citet{chowdhury-2021-modeling}. CRvNN also uses GRC. \textbf{7. BT-GRC:} BT-Cell with GRC cell and plain top-$k$. \textbf{8. BT-GRC + OneSoft:} BT-GRC with OneSoft top-$k$. For experiments with BT-Cell models, we set beam size as $5$ as a practical choice (neither too big nor too small). %However, we also explore beam size $2$ for the most promising variants of BT-Cell to see how far we can get with the minimally costly version of beam search. 

\subsection{ListOps Length Generalization Results} 

\textbf{Dataset Settings:} ListOps \cite{nangia-2018-listops} is a challenging synthetic task that requires solving nested mathematical operations over lists of arguments. We present our results on ListOps in Table \ref{table:listops}. To test for length-generalization performance, we train the models only on sequences with $\leq 100$ lengths (we filter the rest) and test on splits of much larger lengths (eg. $200-300$ or $900-1000$) taken from \citet{havrylov-2019-cooperative}. ``Near-IID" is the original test set of ListOps (it is ``near" IID and not fully IID because a percentage of the split has $> 100$ length sequences whereas such lengths are absent in the training split). We also report the mean accuracy with standard deviation on ListOps in Appendix \ref{mean}. 

\textbf{Results:} \textbf{RecurrentGRC: }As discussed before in $\S$\ref{rnn}, RNNs have to model tree structures implicitly in their bounded hidden states and thus can struggle to generalize to unseen structural depths. This is reflected in the sharp degradation in its length generalization performance. \textbf{GumbelTreeGRCs: } Consistent with prior work \cite{nangia-2018-listops}, Gumbel-Tree models fail to perform well in this task, likely due to their biased gradient estimation. \textbf{CYK-GRC: }CYK-GRC shows some promise to length generalization but it was too slow to run in higher lengths. \textbf{Ordered Memory (OM): } Here, we find that OM struggles to generalize to higher unseen lengths. OM's reliance on soft sequential updates in a nested loop can lead to higher error accumulation over larger unseen lengths or depths.  \textbf{CRvNN: } Consistent with \citet{chowdhury-2021-modeling}, CRvNN performs relatively well at higher lengths. \textbf{BT-GRC: } Here, we find a massive boost over Gumbel-tree baselines even when using the base model. Remarkably, in the $900$-$1000$ length generalization split, BT-GRC increases the performance of GumbelTreeGRC from $37.9\%$ to $86.7\%$---by incorporating beam search with plain top-$k$. \textbf{BT-GRC+OneSoft: } As discussed  in $\S$\ref{topks}, BT-GRC+OneSoft counteracts the bottleneck of gradient propagation being limited through only $k$ beams and achieves near perfect length generalization as we can observe from Table \ref{table:listops}.
\vspace{1mm}

\begin{table*}[t]
\small
\centering
\def\arraystretch{1.2}
\begin{tabular}{  l | l | l l | l l l l l l} 
%\toprule
\hline
\textbf{Model} & \textbf{SST5} & \multicolumn{2}{|c|}{\textbf{IMDB}} & \multicolumn{6}{|c}{\textbf{MNLI}} \\
& IID & Con. OOD & Count. OOD & \textbf{M} & \textbf{MM} & \textbf{Len M} & \textbf{Len MM} & \textbf{Neg M} & \textbf{Neg MM} \\
\hline
RecurrentGRC & $52.19_{1.5}$ & $74.86_{28}$ & $82.72_{19}$ & $71.2_3$ & $71.4_4$ & $49_{25}$ & $49.5_{24}$ & $49.3_6$ & $50.1_6$\\
GumbelTreeGRC & $51.67_{8.8}$ & $70.63_{21}$ & $81.97_5$ & $71.2_7$ & $71.2_6$ & $57.5_{17}$ & $59.6_{12}$ & $50.5_{20}$ & $51.8_{20}$\\
Ordered Memory & $\underline{52.30_{2.7}}$ & $\underline{76.98_{5.8}}$ & $83.68_{7.8}$ & $\mathbf{72.5_3}$ & $\mathbf{73_2}$ & $56.5_{33}$ & $57.1_{31}$ & $50.9_7$ & $51.7_{13}$\\
CRvNN & $51.75_{11}$ & $\mathbf{77.80_{15}}$ & $\underline{85.38_{3.5}}$ & $72.2_4$ & $72.6_5$ & $62_{44}$ & $63.3_{47}$ & $52.8_6$ & $53.8_4$\\
\hline
\multicolumn{5}{l}{Ours}\\
\hline
BT-GRC & $\mathbf{52.32_{4.7}}$ & $75.07_{29}$ & $82.86_{23}$ & $71.6_2$ & $72.3_1$ & $64.7_6$ & $66.4_5$ & $\mathbf{53.7_{37}}$ & $\mathbf{54.8_{43}}$\\
BT-GRC + OneSoft & $51.92_{7.2}$ & $75.68_{21}$ & $84.77_{11}$  & $71.7_1$ & $71.9_2$ & $\mathbf{65.6_{13}}$ & $\mathbf{66.7_9}$ & $53.2_2$ & $54.2_5$\\
\hline
\end{tabular}
%\vspace{+.5em}
\caption{Mean accuracy and standard deviaton on SST5, IMDB, and MNLI. Con. represents Contrast set and Count. represents Countefactuals. Our models were run $3$ times on different seeds. Subscript represents standard deviation. As an example, $90_1 = 90\pm0.1$.}
\label{table:real}
\vspace{-4mm}
\end{table*}

\subsection{ListOps Argument Generalization Results}
\textbf{Dataset Settings:} While length generalization \citep{havrylov-2019-cooperative, chowdhury-2021-modeling} and depth generalization \citep{csordas-2022-ndr} have been tested before for ListOps, the performance on argument generalization is yet to be considered. In this paper, we ask what would happen if we increase the number of arguments in the test set beyond the maximum number encountered during training. The training set of the original ListOps data only has $\leq 5$ arguments for each operator. To test for argument generalization we created two new splits - one with $10$ arguments per operator and another with $15$ arguments per operator. In addition, we also consider the test set of ListOps from Long Range Arena (LRA) dataset \citep{tay2021long} which serves as a check for both length generalization (it has sequences of length $2000$) and argument generalization  (it has $10$ arguments) simultaneously.\footnote{Note that the LRA test set is in-domain for the LRA training set and thus, does not originally test for argument generalization.} The results are in Table \ref{table:listops}.

\textbf{Results:} Interestingly, we find that all the models perform relatively poorly ($<90\%$) on argument generalization. Nevertheless, after OM, BT-GRC-based models perform the best in this split. Comparatively, OM performs quite well in this split - even better than GoldTreeGRC. This shows that the performance of OM is not due to just better parsing. We can also tell that OM's performance is not just for its recursive cell (GRC) because it is shared by other models as well that do not perform nearly as well. This may suggest that the memory-augmented RNN style setup in OM is more amenable for argument generalization. Note that Transformer-based architectures tend to get $\leq 40\%$ on LRA test set for ListOps \citep{tay2021long} despite training on in-distribution data whereas BT-GRC can still generalize to a performance ranging in between $60$-$80\%$ in OOD settings.

%However, with the exception of Ordered Memory (OM), BT-Cell models perform much better than any other models (including otherwise strong contenders like CRvNN). Interestingly,  Some potential reasons could be the following; First, we know OM's performance is not simply due to better parsing because it even surpasses GoldTreeGRC at times. Second, we also know OM's performance is not just due to a better recursive cell, since its cell (GRC) is shared by many other models that do not perform as well. This may suggest that the memory-augmented RNN style setup in OM is more amenable for argument generalization. Note that Transformer-based architectures tend to get $\leq 40\%$ on LRA test set for ListOps \citep{tay2021long} despite training on in-distribution data whereas we can still generalize to a performance ranging in between $70$-$80\%$ by using RvNN-based models trained on data with much limited sequence lengths and fewer arguments. 

\subsection{Semantic Analysis (SST and IMDB) Results} 

\textbf{Dataset Settings: } SST5 \citep{socher-etal-2013-recursive} and IMDB \citep{maas-etal-2011-learning} are natural language classification datasets (for sentiment classification). For IMDB, to focus on OOD performance, we also test our models on the contrast set (Con.) from \cite{gardner-etal-2020-evaluating} and the counterfactual test set (Count.) from \cite{Kaushik2020Learning}. We present our results on these datasets in Table \ref{table:real}.

\textbf{Results:} The results in these natural language tasks are rather mixed. There are, however, some interesting highlights. CRvNN and OM do particularly well in the OOD splits (contrast set and counterfactual split) of IMDB, correlating with their better OOD generalization in synthetic data. BT-GRC + OneSoft remains competitive in those splits with OM and CRvNN and is better than any other models besides CRvNN and OM. STE Gumbel-based models tend to perform particularly worse on IMDB.

\subsection{Natural Language Inference Experiments}
\label{nli}
\textbf{Dataset Settings:} We ran our models on MNLI \citep{william2018broad} which is a natural language inference task. We tested our models on the development set of MNLI and used a randomly sampled subset of $10,000$ data points from the original training set as the development set.  Our training setup is different from \citet{chowdhury-2021-modeling} and other prior latent tree models that combine SNLI \citep{bowman-etal-2015-large} and MNLI training sets (in that we do not add SNLI data to MNLI). We filter sequences $\geq 150$ length from the training set for efficiency. We also test our models in various stress tests \citep{naik-etal-2018-stress}. We report the results in Table \ref{table:real}. In the table, M denotes matched development set (used as test set) of MNLI. MM denotes mismatched development set (used as test set) of MNLI. LenM, LenMM, NegM, and NegMM denote Length Match, Length Mismatch, Negation Match, and Negation Mismatch stress test sets, respectively - all from \citet{naik-etal-2018-stress}. Len M/Len MM add to the length by adding tautologies. Neg M/Neg MM add tautologies containing ``not" terms which can bias the model to falsely predict contradictions. 

%LenM denotes length matched stress set from \citep{naik-etal-2018-stress}. LenMM denotes length mismatched stress set from \citep{naik-etal-2018-stress}. NegM denotes negation matched stress set from \citep{naik-etal-2018-stress}. NegMM denotes negation mismatched stress set from \citep{naik-etal-2018-stress}. Len M/Len MM stress sets add to the length of the premise by adding tautologies. Neg M/Neg MM stress sets add tautologies containing ``not" terms which can bias the model to falsely predict contradictions. 

\textbf{Results:} The results in Table \ref{table:real} show that BT-GRC models perform comparably with the other models in the standard matched/mismatched sets (M and MM). However, they outperform all the other models on Len M and Len MM. Also, BT-GRC models tend to do better than the other models in Neg M and Neg MM. Overall, BT-Cell shows better robustness to stress tests. 
\begin{table*}[t]
\small
\centering
\def\arraystretch{1.2}
\begin{tabular}{  l | l | l l l | l l |l} 
%\toprule
\hline
\textbf{Model} & \textbf{near-IID} & \multicolumn{3}{c}{\textbf{Length Gen.}} & \multicolumn{2}{|c|}{\textbf{Argument Gen.}} & \textbf{LRA}\\
(Lengths) & $\leq$ 1000 & 200-300 & 500-600 & 900-1000 & 100-1000 & 100-1000 & $2000$\\
(Arguments) & $\leq$ 5 & $\leq$ 5 & $\leq$ 5 & $\leq$ 5 & 10 & 15 & 10\\
\hline
\multicolumn{8}{l}{BT-GRC Models (Beam size $5$)}\\
\hline
BT-GRC & $99.39$ & $96.15$ & $92.55$ & $86.7$ & $\mathbf{77.1}$ & $63.7$ & $67.3$\\
BT-GRC + OneSoft & $\mathbf{99.92}$ & $\mathbf{99.5}$ & $\mathbf{99}$ & $\mathbf{97.2}$ & $76.05$ & $\mathbf{67.9}$ & $\mathbf{71.8}$\\
\hline
\multicolumn{8}{l}{Alternative models in the Vicinity of BT-GRC}\\
\hline
BT-LSTM & $94.1$ & $85.1$ & $83.5$ & $78.8$ & $67.9$ & $44.3$ & $57.9$\\
BSRP-GRC & $70.3$ & $42.4$ & $33.2$ & $26.3$ & $40.2$ & $35.8$ & $29.7$\\
MC-GumbelTreeGRC & $89.3$ & $36.8$ & $28.2$ & $25.1$  & $39.5$ & $34$ & $30.1$\\
BT-GRC+SOFT & $69$ & $44$ & $37.1$ & $29.4$  &  $39.5$ & $38.6$ & $31.6$\\
\hline
\multicolumn{8}{l}{Robustness of OneSoft Top-K to lower Beam Size (size $2$)}\\
\hline
BT-GRC & $94.18$ & $68.2$ & $56.85$ & $50.2$ & $64.45$ & $56.95$ & $55.85$\\
BT-GRC + OneSoft & $99.69$ & $97.55$ & $95.40$ & $91$ & $75.75$ & $62$ & $66.1$\\
\hline
\end{tabular}
\caption{Accuracy of different models on ListOps (same setting as in Table \ref{table:listops}). We report the median of $3$ runs.}
\label{table:listopsalter}
\end{table*}
\section{Analysis}
\subsection{Analysis of Neighbor Models}
\label{ablations}
We also analyze some other models that are similar to BT-GRC in Table \ref{table:listopsalter} as a form of ablation and show that BT-GRC is still superior to them. We describe these models below and discuss their performance on ListOps. 

\textbf{BT-LSTM:} This is just BT-Cell with an LSTM cell \cite{hochreiter1997long} instead of GRC. In Table \ref{table:listopsalter}, we find that BT-LSTM can still perform moderately well (showing the robustness of BT-Cell as a framework) but worse than what it can do with GRC. This is consistent with prior works showing superiority of GRC as a cell function \cite{shen-2019-ordered,chowdhury-2021-modeling}.

\textbf{BSRP-GRC:} This is an implementation of Beam Shift Reduce Parser \cite{maillard-clark-2018-latent} with a GRC cell. Similar to us, this approach applies beam search but to a shift-reduce parsing model as elaborated in Appendix \ref{BSRP}. Surprisingly, despite using a similar framework to BT-Cell, BSRPC-GRC performs quite poorly in Table \ref{table:listopsalter}. We suspect this is because of the limited gradient signals from its top-$k$ operators coupled with the high recurrent depth for backpropagation (twice the sequence length) encountered in BSRP-GRC compared to that in BT-Cell (the recurrent depth is the tree depth). Moreover, BSRP-GRC, unlike BT-Cell, also lacks the global competition among all parent compositions when making shift/reduce choices.

\textbf{MC-GambelTreeGRC:} Here we propose a Monte Carlo approach towards Gumbel Tree GRC. This model runs $k$ gumbel-tree models with shared parameters in parallel. Since the models are stochastic, they can sample different latent structures. In the end we can average the final $k$ sentence encodings treating this as a Monte-Carlo approximation. We set $k=5$ to be comparable with BT-Cell. MC-GumbelTreeGRC is similar to BT-Cell because it can model different structural interpretations. However, it fails to do as effectively as BT-Cell in ListOps. We suspect this is because beam-search based structure selection allows more competition between structure candidates when using top-$k$ for truncation and thus enables better structure induction.

\textbf{BT-GRC+SOFT:} This model incorporates another potential alternative to OneSoft within BT-GRC. It uses a differentiable sorting algorithm, SOFT Top-$k$, that was previously used in beam search for language generation \cite{xie-2020-differentiable}, to implement the top-$k$ operator replacing OneSoft. However, it performs poorly. Its poor performance supports our prior conjecture ($\S$\ref{topks}) that using a soft permutation matrix in all recursive iterations is not ideal because of increased chances of error accumulation and more “washing out” through weighted averaging of distinct beams.

\subsection{OneSoft Top-$k$ with Lower Beam Size}
We motivated ($\S$\ref{topks}) the proposal of OneSoft top-$k$ to specifically counteract the bottleneck of gradient propagation being limited through only $k$ beams in the base BT-Cell model (with plain top-$k$). While we validate this bottleneck through our experiments in Table \ref{table:listops} for beam size 5, the bottleneck should be even worse when $k$ (beam size) is low (e.g., $2$). Based on our motivation, OneSoft should perform much better than plain top-$k$ when beam size is low. We perform experiments with beam size $2$ on ListOps to understand if that is true and show the results in Table \ref{table:listopsalter}. As we can see, OneSoft indeed performs much better than plain top-$k$ with lower beam size of 2 where BT-GRC gets only $50.2\%$ in the $900$-$1000$ split of ListOps, and BT-GRC+OneSoft gets $91\%$. As we would expect beam size $5$ (from Table \ref{table:listops} and also shown verbatim in Table \ref{table:listopsalter}) still outperforms beam size $2$ in a comparable setting. We report some additional results with beam size $2$ in Appendix \ref{real2}. 

\begin{table*}[t]
\small
\centering
\def\arraystretch{1.2}
\begin{tabular}{  l | l l | l l | l l} 
%\toprule
\hline
& \multicolumn{6}{c}{\textbf{Sequence Lengths}}\\ 
\hline
\textbf{Model} & \multicolumn{2}{c|}{$\mathbf{200-250}$} & \multicolumn{2}{c|}{$\mathbf{500-600}$} & \multicolumn{2}{c}{$\mathbf{900-1000}$} \\
& Time & Memory & Time & Memory & Time & Memory \\
\hline
RecurrentGRC & $0.2$ min & $0.02$ GB & $0.5$ min & $0.02$ GB & $1.3$ min & $0.03$ GB\\
GumbelTreeGRC  & $0.5$ min & $0.35$ GB & $2.1$ min & $1.95$ GB & $3.5$ min & $5.45$ GB\\
CYK-GRC & $9.3$ min & $32.4$ GB & OOM & OOM & OOM & OOM\\
BSRP-GRC & $2.3$ min & $0.06$ GB & $6.1$ min & $0.19$ GB & $10.5$ min & $0.42$ GB\\
Ordered Memory & $8.0$ min & $0.09$ GB & $20.6$ min & $0.21$ GB & $38.2$ min & $0.35$ GB \\
CRvNN & $1.5$ min & $1.57$ GB & $4.3$ min & $12.2$ GB & $8.0$ min & $42.79$ GB\\
MC-GumbelTreeGRC & $1.1$ min & $1.71$ GB & $2.4$ min & $9.85$ GB & $4.3$ min & $27.33$ GB\\
BT-GRC & $1.1$ min & $1.71$ GB & $2.6$ min & $9.82$ GB & $5.1$ min & $27.27$ GB\\
BT-GRC + OneSoft  & $1.4$ min & $2.74$ GB & $4.0$ min & $15.5$ GB & $7.1$ min & $42.95$ GB\\
BT-GRC + SOFT  & $5.1$ min & $2.67$ GB & $12.6$ min & $15.4$ GB & $23.1$ min & $42.78$ GB \\
\hline
\end{tabular}
%\vspace{+.5em}
\caption{Empirical time and memory consumption for various models on an RTX A6000. Ran on $100$ ListOps data with batch size $1$}.
\label{table:speedtest}
\vspace{-4mm}
\end{table*}
\subsection{Efficiency Analysis}
\label{efficiency}
\textbf{Settings:} In Table \ref{table:speedtest}, we compare the empirical performance of various models in terms of time and memory. We train each model on ListOps splits of different sequence lengths ($200$-$250$, $500$-$600$, and $900$-$1000$). Each split contains $100$ samples. Batch size is set as $1$. Other hyperparameters are the same as those used for ListOps. For CRvNN, we show the worst case performance (without early halt) because otherwise it halts too early without learning to halt from more training steps or data. 

\textbf{Discussion:} RecurrentGRC and GumbelTreeGRC can be relatively efficient in terms of both runtime and memory consumption. BSRP-GRC and OM, being recurrent models, can be highly efficient in terms of memory but their complex recurrent operations make them slow. CYK-GRC is the worst in terms of efficiency because of its expensive chart-based operation. CRvNN is faster than OM/BSRP-GRC but its memory consumption can scale worse than BT-GRC because of Transformer-like attention matrices for neighbor retrieval. MC-GumbelTreeGRC is similar to a batched version of GumbelTreeGRC. BT-GRC performs similarly to MC-GumbelTreeGRC showing that the cost of BT-GRC is similar to increasing batch size of GumbelTreeGRC. BT-GRC + OneSoft perform similarly to CRvNN. BT-GRC + SOFT is much slower due to using a more expensive optimal transport based differentiable sorting mechanism (SOFT top-$k$) in every iteration. This shows another advantage of using OneSoft over other more sophisticated alternatives.  

\subsection{Additional Analysis and Experiments}

\textbf{Heuristics Tree Models:} We analyze heuristics-based tree models (Random tree, balanced tree) in Appendix \ref{heuristics}. 

\textbf{Synthetic Logical Inference}: We present our results on a challenging synthetic logical inference task \citep{bowman-2015-tree} in Appendix \ref{synthlogic}. We find that most variants of BT-Cell can perform on par with prior SOTA models. 

\textbf{Depth Generalization}: We also run experiments to test depth-generalization performance on ListOps (Appendix \ref{dg_exp}). We find that BT-Cell can easily generalize to much higher depths and it does so more stably than OM.

\textbf{Transformers}: We experiment briefly with Neural Data Routers  \citep{csordas-2022-ndr} which is a  Transformer-based model proven to do well in tasks like ListOps. However, we find that Neural Data Routers (NDRs), despite their careful inductive biases, still struggle with sample efficiency and length generalization compared to strong RvNN-based models. We discuss more in Appendix \ref{ndr_exp}. 

\textbf{Parse Tree Analysis}: We analyze parsed trees and score distributions in Appendix \ref{parse_analysis}.

\section{Related Works}
\citet{goldberg-2010-efficient} proposed the easy-first algorithm for dependency parsing. \citet{ma-etal-2013-easy} extended it with beam search for parsing tasks. \citet{choi-2018-learning} integrated easy-first-parsing with an RvNN. Similar to us, \citet{maillard-clark-2018-latent} used beam search to extend shift-reduce parsing whereas \citet{drozdov-etal-2020-unsupervised} used beam search to extend CYK-based algorithms. However, BT-Cell-based models achieve higher accuracy than the former style of models (e.g., BSRP-GRC) and are computationally more efficient than the latter style of models (e.g., CYK-GRC). Similar to us, \citet{collobert-2019-fully-differentiable} also used beam search in an end-to-end fashion during training but in the context of sequence generation. However, none of the above approaches explored beyond hard top-$k$ operators in beam search. One exception is \citet{xie-2020-differentiable} where a differentiable top-$k$ operator (SOFT Top-$k$) is used in beam search for language generation but as we show in $\S$\ref{ablations} it does not work as well. Another exception is \citet{goyal2018continuous} where an iterated softmax is used to implement a soft top-$k$ operator for differentiable beam search. However, iterated softmax operations can slow down BT-GRC and overall share similar limitations as SOFT Top-$k$. Moreover, SOFT Top-$k$ was shown to perform slightly better than iterated softmax \cite{xie-2020-differentiable} and we show that our OneSoft fares better than SOFT Top-$k$ in $\S$\ref{ablations} for our contexts.  Besides \citet{xie-2020-differentiable}, there are multiple versions of differentiable top-k operators or sorting functions \citep{adams-2011-ranking,Ploetz:2018:NNN,grover2018stochastic,cuturi-2019-differentiable,xie-2020-differentiable,blondel2020fast, peterson-2021-differentiable, petersen2022monotonic} (interalia). We leave a more exhaustive analysis of them as future work. However, note that some of them would suffer from the same limitations as SOFT top-k \cite{xie-2020-differentiable} - that is, they can significantly slow down the model and they can lead to ``washed out" representations. We provide an extended related work survey in Appendix \ref{ext_related_works}.

\section{Discussion}
In this section, we first discuss the trade offs associated with different RvNN models that we compare. We then highlight some of the features of our BT-Cell model. 

\textbf{CYK-Cell:} Our experiments do not show any empirical advantage of CYK-Cell compared to CRvNN/OM/BT-Cell. Moreover, computationally it offers the worst trade-offs. However, there are some specialized ways \cite{drozdov-etal-2019-unsupervised-latent,drozdov-etal-2020-unsupervised} in which CYK-Cell-based models can be used for masked language modeling that other models cannot. Furthermore, \citet{hu-etal-2021-r2d2,hu-etal-2022-fast} also propose several strategies to make them more efficient in practice. %Nevertheless, There is room for further exploration and development of the other models - CRvNN/OM/BT-Cell. One under-explored area would be providing top-down signals to contextualize token representations based on high-level span representations \cite{teng-zhang-2017-head}. 

\textbf{Ordered Memory (OM)}: OM is preferable when memory is a priority over time. Its low memory also allows for high batch size which alleviates its temporal cost. OM shows some length generalization issues but overall performs well in general. It can also be used for autoregressive language modeling in a straightforward manner.

\textbf{CRvNN:} CRvNN also generally performs competitively. It can be relatively fast with dynamic halting but its memory complexity can be a bottleneck; although, that can be mitigated by fixing an upper-bound to maximum recursion.

\textbf{BT-Cell Features:} We highlight the salient features of BT-Cell below:
\begin{enumerate}
    \item BT-Cell's memory consumption is better than CRvNN (without halt) but its speed is generally slower than CRvNN (but faster than Ordered Memory).
    \item BT-Cell as a framework can be easier to build upon for its conceptual simplicity than OM/CRvNN/CYK-Cell. 
    \item Unlike CRvNN and OM, BT-Cell also provides all the intermediate node representations (spans) of the induced tree. Span representations can often have interesting use cases - they have been used in machine translation \cite{Su2020neural,patel2022forming}, for enhancing compositional generalization \cite{bogin-etal-2021-latent,herzig-berant-2021-span}, or other natural language tasks \cite{patel2022forming} in general. We leave possible ways of integrating BT-Cell with other deep learning modules as a future work. BT-Cell can also be a drop-in replacement for Gumbel Tree LSTM \cite{choi-2018-learning}. 
    \item With BT-Cell, we can extract tree structures which can offer some interpretability. The extracted structures can show some elements of ambiguities in parsing (different beams can correspond to different interpretations of ambiguous sentences). See Appendix \ref{parse_analysis} for more details on this. 
\end{enumerate}

We also note that OneSoft, on its own, can be worthy of individual exploration as a semi-differentiable top-$k$ function. Our experiments show comparative advantage of it over a more sophisticated optimal-transport based method for implementation of differentiable top-$k$ (SOFT top-$k$) \cite{xie-2020-differentiable}. In principle, OneSoft can serve as a general purpose option whenever we need differentiable top-$k$ selection in neural networks.

\section{Limitations}
While our Beam Tree Cell can serve as a ``middle way'' between Gumbel Tree models and CYK-based models in terms of computational efficiency, the model is still quite expensive to run compared to even basic RNNs. Moreever, the study in this paper is only done in a small scale setting without pre-trained models and only in a single natural language (English). More investigation needs to be done in the future to test for cross-lingual modeling capacities of these RvNN models and for ways to integrate them with more powerful pre-trained architectures.

\section{Conclusion}
We present BT-Cell as an intuitive way to extend RvNN that is nevertheless highly competitive with more sophisticated models like Ordered Memory (OM) and CRvNN. In fact, BT-Cell is the only model that achieves moderate performance in argument generalization while also excelling in length generalization in ListOps. It also shows more robustness in MNLI, and overall it is much faster than OM or CYK-GRC. We summarize our main results in Appendix \ref{main_summary}. The ideal future direction would be to focus on argument generalization and systematicity while maintaining computational efficiency. We also aim for added flexibility for handling more relaxed structures like non-projective trees or directed acyclic graphs as well as richer classes of languages \citep{dusell2022learning, Deletang2022NeuralNA}. 
%Overall, we find all three of Ordered Memory (OM), CRvNN, and BT-Cell are competitive against each other; none being completely superior in all aspects. BT-Cell with OneSoft excels in length generalization at ListOps and offers moderate performance on argument generalization. OM excels in argument generalization while struggling in length generalization. CRvNN performs decently in length generalization but struggles in argument generalization. CYK-GRC shows some promise too but it is several times more expensive to run while having poor argument generalization. We discuss some limitations to address in future works in Appendix \ref{limits}.

% In the unusual situation where you want a paper to appear in the
% references without citing it in the main text, use \nocite
%\nocite{langley00}

\section*{Acknowledgments}
This research is supported in part by NSF CAREER award \#1802358, NSF IIS award \#2107518, and UIC Discovery Partners Institute (DPI) award. Any opinions, findings, and conclusions expressed here are those of the authors and do not necessarily reflect the views of NSF or DPI. We thank our anonymous reviewers for their constructive feedback. 

\bibliography{main}
\bibliographystyle{icml2023}

%%%%%%%%%%%%%%%%%%%%%%%%%%%%%%%%%%%%%%%%%%%%%%%%%%%%%%%%%%%%%%%%%%%%%%%%%%%%%%%
%%%%%%%%%%%%%%%%%%%%%%%%%%%%%%%%%%%%%%%%%%%%%%%%%%%%%%%%%%%%%%%%%%%%%%%%%%%%%%%
% APPENDIX
%%%%%%%%%%%%%%%%%%%%%%%%%%%%%%%%%%%%%%%%%%%%%%%%%%%%%%%%%%%%%%%%%%%%%%%%%%%%%%%
%%%%%%%%%%%%%%%%%%%%%%%%%%%%%%%%%%%%%%%%%%%%%%%%%%%%%%%%%%%%%%%%%%%%%%%%%%%%%%%
\newpage
\appendix
\onecolumn
\begin{algorithm}[t]
   \caption{Easy First Composition}
   \label{alg:easy-first}
\begin{algorithmic}
   \STATE {\bfseries Input:} data $X = [x_1, x_2,....x_n]$
   \WHILE{True}
    \IF{$len(X)==1$}
    \STATE return $X[0]$
    \ENDIF
    \IF{$len(X)==2$}
    \STATE return $cell(X[0], X[1])$
    \ENDIF
   \STATE $Children_L$, $Children_R$ $\gets$ $X[:len(X)-1],X[1:]$
   \STATE $Parents$ $\gets$ [$cell(child_L, child_R)$ for $child_L$, $child_R$ in $zip(Children_L, Children_R]$
   \STATE $Scores \gets [scorer(parent)\text{ for }parent\text{ in }Parents]$
   \STATE $index$ $\gets$ $argmax(Scores)$
   \STATE $X[index] \gets Parents[index]$
   \STATE Delete $X[index+1]$
   \ENDWHILE
\end{algorithmic}
\end{algorithm}

\begin{algorithm}[t]
   \caption{Beam Tree Cell}
   \label{alg:bt-cell}
\begin{algorithmic}
   \STATE {\bfseries Input:} data $X = [x_1, x_2,....x_n], k\text{ (beam size)}$
   \STATE $BeamX$ $\gets$ [$X$]
   \STATE $BeamScores$ $\gets$ [0]
   \WHILE{True}
    \IF{$len(BeamX[0])==1$}
    \STATE $BeamX$ $\gets$ [$beam[0]$ for $beam$ in $BeamX$]
    \STATE break
    \ENDIF
    \IF{$len(BeamX[0])==2$}
    \STATE $BeamX$ $\gets$ [$cell(beam[0], beam[1])$ for $beam$ in $BeamX$]
    \STATE break
    \ENDIF
   \STATE $NewBeamX$ $\gets$ $[]$
   \STATE $NewBeamScores$ $\gets$ $[]$
   \STATE \FOR{$Beam$,$BeamScore$ in $zip(BeamX, BeamScores)$}
   \STATE $Parents \gets [cell(beam[i], beam[i+1])\text{ for }i\text{ in }range(0,len(beam)-1)]$
   \STATE $Scores \gets log \circ softmax([scorer(parent)\text{ for }parent\text{ in }Parents])$
   \STATE $Indices \gets topk(Scores, k)$
   \STATE \FOR{$i$ in $range(K)$}
   \STATE $newBeam \gets$ $deepcopy(Beam)$
   \STATE $newBeam[Indices[i]] \gets Parents[Indices[i]]$
   \STATE Delete $newBeam[Indices[i]+1]$
   \STATE $NewBeamX.append(newBeam)$
   \STATE $newScore \gets BeamScore + Scores[indices[i]]$
   \STATE $newBeamScores.append(newScore)$
   
   \ENDFOR
   \ENDFOR
   \STATE $Indices \gets topk(newBeamScores, k)$
   \STATE $BeamScores \gets [newBeamScores[i]\text{ for }i\text{ in Indices}]$
   \STATE $BeamX \gets [newBeamX[i]\text{ for }i\text{ in Indices}]$
   \ENDWHILE
   \STATE $BeamScores \gets Softmax(BeamScores)$
   \STATE Return $sum([score * X \text{ for }score,X\text{ in }zip(BeanScores,BeamX)])$
\end{algorithmic}
\end{algorithm}

\section{Pseudocodes}
We present the pseudocode of the easy first composition in Algorithm \ref{alg:easy-first} and the pseudocode of BT-cell in Algorithm \ref{alg:bt-cell}. Note that the algorithms are written as they are for the sake of illustration: in practice, many of the nested loops are made parallel through batched operations in GPU (Model code is available in github: \url{https://github.com/JRC1995/BeamTreeRecursiveCells/blob/main/models/layers/BeamGumbelTreeCell.py}). 

\subsection{Beam Tree Cell Algorithm}
\label{btcell-algo}
Here, we briefly describe the algorithm of BT-cell (Algorithm \ref{alg:bt-cell}) in words. 
In BT-Cell, instead of maintaining a single sequence per sample, we maintain some $k$ (initially $1$) number of sequences and their corresponding scores (initialized to $0$). $k$ is a hyperparameter defining the beam size. Each sequence (henceforth, interchangeably referred to as ``beam") is a hypothesis representing a particular sequence of choices of parents. Thus, each beam represents a different path of composition (for visualization see Figure \ref{fig:softpath}). At any moment the score represents the log-probability for its corresponding beam. The steps in each iteration of the recursion of BT-Cell are as follows: \textbf{Step 1:} similar to gumbel-tree models, we create all candidate parent compositions for each of the $k$ beams. \textbf{Step 2:} we score the candidates with the $score$ function (defined in $\S$\ref{easy-first}). \textbf{Step 3:} we choose top-$k$ highest scoring candidates. We treat the top-$k$ choices as mutually exclusive. Thus, each of the $k$ beams encounters $k$ branching choices, and are updated into $k$ distinct beams (similar to before, the children are replaced by the chosen parent). Thus, we get $k \times k$ beams. \textbf{Step 4:} we update the beam scores. The sub-steps involved in the update are described next. \textbf{Step 4.1: } we apply a log-softmax to the scores of the latest candidates to put the scores into the log-probability space. \textbf{Step 4.2:} we add the log-softmaxed scores of the latest chosen candidate to the existing beam score for the corresponding beam where the candidate is chosen. As a result, we will have $k \times k$ beam scores. \textbf{Step 5:} we truncate the $k \times k$ beams and beam scores into $k$ beams and their corresponding $k$ scores to prevent exponential increase of the number of beams. For that, we again simply use a top-$k$ operator to keep only the highest scored beams. 

At the end of the recursion, instead of a single item representing the sequence-encoding, we will have $k$ beams of items with their $k$ scores. At this point, to get a single item, we do a weighted summation with the softmaxed scores as the weights as described in $\S$\ref{bt-cell}.

Note that the current method of beam search does not necessarily return unique beams. They do return unique sequences of parsing actions but different sequence of parsing actions can end up corresponding to the same structure. We leave it for future exploration to investigate efficient ways to restrict duplicates and check whether that helps.  

\section{Gated Recursive Cell (GRC)}
The Gated Recursive Cell (GRC) was originally introduced by \cite{shen-2019-ordered} drawing inspiration from the Transformer's feed-forward networks. In our implementation, we use the same variant of GRC as was used in \cite{chowdhury-2021-modeling} where a GELU \cite{hendrycks2016gelu} activation function was used. 
\label{grc}
We present the equations of GRC here:
\begin{equation}
    \left[
    \begin{matrix}
        z_i \\ 
        h_i \\
        c_i \\
        u_i
    \end{matrix}
    \right]=   \mathrm{GeLU} \left(\left[ 
    \begin{matrix}
        child_{left} \\ 
        child_{right}
    \end{matrix} 
    \right]W^{Cell}_1 
    + b_1 \right)W^{cell}_2 + b_2
\end{equation}
\begin{eqnarray}
o_i = LN (\sigma(z_i) \odot child_{left}
                    + \sigma(h_i) \odot child_{right} 
                    + \sigma(c_i) \odot u_i) \label{eq:gated_sum}
\end{eqnarray}
$\sigma$ is $sigmoid$; $o_i$ is the parent composition $\in {\rm I\!R}^{d_h}$; $child_{left}, child_{right} \in {\rm I\!R}^{d_h}$;
$W_1^{cell} \in {\rm I\!R}^{2d_h \times 4d_h}$; $b_1 \in {\rm I\!R}^{4d_h}$; $W_2 \in {\rm I\!R}^{4d_{h} \times d_h}$; $b_1 \in {\rm I\!R}^{d_{h}}$. We use this same GRC function for any recursive model (including our implementation of Ordered Memory) that constitutes GRC.

\section{BSRP-GRC Details}
\label{BSRP}
For the decisions about whether to {\em shift} or {\em reduce}, we use a scorer function similar to that used in \cite{chowdhury-2021-modeling}. While \cite{chowdhury-2021-modeling} use the decision function on the concatenation of local hidden states ($n$-gram window), we use the decision function on the concatenation of the last two items in the stack and the next item in the queue. The output is a scalar sigmoid activated logit score $s$. We then treat $log(s)$ as the score for {\em reducing} in that step, and $log(1-s)$ as the score for {\em shifting} in that step. The scores are manipulated appropriately for edge cases (when there are no next item to shift, or when there are no two items in the stack to reduce). Besides that, we use the familiar beam search strategy over standard shift-reduce parsing. Finally, the beams of final states are merged through the weighted summation of the states based on the softmaxed scores of each beam similar to the BT-Cell model as described in $\S$\ref{bt-cell}. 

\section{Results Summary}
\label{main_summary}
In this section, we summarize our main findings throughout the paper (appendix included): 

\begin{enumerate}
    \item In ListOps, BT-GRC + OneSoft shows near-perfect length generalization (Table \ref{table:listops}) and near-perfect depth generalization performance (Table \ref{table:listopsdg}). Ordered Memory \cite{shen-2019-ordered}, which is otherwise a strong contender, can fall behind in this regard. Even Neural Data Router \cite{csordas-2022-ndr} (which is a Transformer-based model with special inductive biases) still struggles when trying to generalize to depths/lengths several times higher than what was seen in the training set  ($\S$ \ref{ndr_exp}).
    \item We show that argument generalization (previously never investigated) in ListOps is still a challenge and remains unsolved by RvNNs even with ground truth trees. Among the models without ground truth trees, Ordered Memory shows the best results and BT-Cell variants show the second-best results (Table \ref{table:listops}).
    \item BT-Cell keeps up with SOTA in a challenging synthetic logical inference task, whereas models like GumbelTreeGRC and RecurrentGRC fall behind (Table \ref{table:PNLIRESULTS}).
    \item BT-Cell keeps up with the other RvNN models in natural language tasks like sentiment classification and natural language inference. It is worth noting that it does particularly better in the stress tests of MNLI compared to other existing RvNN models (Table \ref{table:real}).
\end{enumerate}

\section{Additional Experiments and Analysis}

\begin{table*}[t]
\small
\centering
\def\arraystretch{1.2}
\begin{tabular}{  l | l | l l l | l l |l} 
%\toprule
\hline
\textbf{Model} & \textbf{DG} & \multicolumn{3}{c}{\textbf{Length Gen.}} & \multicolumn{2}{|c|}{\textbf{Argument Gen.}} & \textbf{LRA}\\
(Lengths) & $\leq$ 100 & 200-300 & 500-600 & 900-1000 & 100-1000 & 100-1000 & 2000\\
(Arguments) & $\leq$ 5 & $\leq$ 5 & $\leq$ 5 & $\leq$ 5 & 10 & 15 & $\leq$ 10\\
(Depths) & 8-10 & $\leq$ 20 & $\leq$ 20 & $\leq$ 20 & $\leq$ 10 & $\leq$ 10 & $\leq$ 10\\
\hline
\multicolumn{5}{l}{\textit{With gold trees}}\\
\hline
GoldTreeGRC & $99.95$ & $99.95$ & $99.9$ & $99.8$ & $76.95$ & $77.1$ & $74.55$\\
\hline
\multicolumn{5}{l}{\textit{Baselines without gold trees}}\\
\hline
CYK-GRC & $99.45$ & $99.0$ & --- & --- & $67.8$ & $35.15$ & ---\\
Ordered Memory & $\mathbf{99.95}$ & $\underline{99.8}$ & $99.25$ & $96.4$ & $\mathbf{79.95}$ & $\mathbf{77.55}$ & $\mathbf{77}$\\
CRvNN & $\underline{99.9}$ & $99.4$ & $99.45$ & $98.9$ & $65.7$ & $43.4$ & $65.1$\\
\hline
\multicolumn{5}{l}{Ours}\\
\hline
BT-GRC & $\mathbf{99.95}$ & $\mathbf{99.95}$ & $\mathbf{99.95}$ & $\mathbf{99.9}$ & $75.35$ & $72.05$ & $68.1$\\
BT-GRC + OneSoft & $\underline{99.9}$ & $99.6$ & $98.1$ & $97.1$ & $\underline{78.1}$ & $71.25$ & $\underline{75.45}$ \\
\hline
\end{tabular}
%\vspace{+.5em}
\caption{Accuracy on ListOps-DG. We report the median of $3$ runs except for CYK-GRC which was run only once for its high resource demands. Our models were trained on lengths $\leq$ 100, depth $\leq$ 6, and arguments $\leq$ 5. We bold the best results and underline the second-best among models that do not use gold trees.}.
\label{table:listopsdg}
\vspace{-4mm}
\end{table*}

\subsection{ListOps-DG Experiment}
\label{dg_exp}
\textbf{Dataset Settings: }The length generalization experiments in ListOps do not give us an exact perspective in depth generalization\footnote{By depth, we simply mean the maximum number of nested operators in a given sequence in the case of ListOps.} capacities. So there is a question of how models will perform in unseen depths. To check for this, we create a new ListOps split which we call ``ListOps-DG". For this split, we create $100,000$ training data with arguments $\leq 5$, lengths $\leq 100$, and depths $\leq 6$. We create $2000$ development data with arguments $\leq 5$, lengths $\leq 100$, and depths $7$. We create $2000$ test data with arguments $\leq 5$, lengths $\leq 100$, and depths $8$-$10$. In addition, we tested on the same length-generalization splits as before from \citet{havrylov-2019-cooperative}. Those splits have a maximum of $20$ depth; thus, they can simultaneously test for both length generalization and depth generalization capacity.  We also use the argument generalization splits, and LRA test split as before. The results are presented in Table \ref{table:listopsdg}. We only evaluate the models that were promising ($\geq 90\%$ in near IID settings) in the original ListOps split. We report the median of $3$ runs for each model except CYK-GRC which was too expensive to run (so we ran once). 
%(which now simultaneously have much higher depths too: $\leq 20$),

\textbf{Results: }Interestingly, we find that base BT-GRC, CRvNN, and Ordered Memory now do much better in length generalization compared to the original ListOps split. We think this is because of the increased data (the training data in the original ListOps is $\sim 75,000$ after filtering data of length $>100$ whereas here we generated $100,000$ training data). However, while the median of $3$ runs in Ordered Memory is decent, we found one run to have very poor length generalization performance. To investigate more deeply if Ordered Memory has a particular stability issue, we ran Ordered Memory for $10$ times with different seeds, and we find that it frequently fails to learn to generalize over length. As a baseline, we also ran BT-GRC similarly for $10$ runs and found it to be much more stable.  We report the mean and standard deviation of $10$ runs of Ordered Memory and BT-GRC in Table \ref{table:listopsdg_stability}. As can be seen, the mean of BT-GRC is much higher than that of Ordered Memory in length generalization splits.

\begin{table*}[t]
\small
\centering
\def\arraystretch{1.2}
\begin{tabular}{  l | l | l l l | l l |l} 
%\toprule
\hline
\textbf{Model} & \textbf{DG} & \multicolumn{3}{c}{\textbf{Length Gen.}} & \multicolumn{2}{|c|}{\textbf{Argument Gen.}} & \textbf{LRA}\\
(Lengths) & $\leq$ 100 & 200-300 & 500-600 & 900-1000 & 100-1000 & 100-1000 & 2000\\
(Arguments) & $\leq$ 5 & $\leq$ 5 & $\leq$ 5 & $\leq$ 5 & 10 & 15 & $\leq$ 10\\
(Depths) & 8-10 & $\leq$ 20 & $\leq$ 20 & $\leq$ 20 & $\leq$ 10 & $\leq$ 10 & $\leq$ 10\\
\hline
\multicolumn{5}{l}{\textit{Stability Test: Mean/Std with 10 runs. Beam size 5 for BT-GRC}}\\
\hline
Ordered Memory & $99.94_{0.6}$ & $97.58_{32}$ & $78.785_{197}$ & $61.85_{291}$ & $77.66_{30}$ & $69.03_{107}$ & $67.35_{125}$\\
BT-GRC & $99.84_{1.5}$ & $99.58_{5.8}$ & $98.8_{21}$ & $97.85_{39}$ & $73.82_{57}$ & $ 66.21_{107}$ & $66.975_{102}$\\
\hline
\end{tabular}
%\vspace{+.5em}
\caption{Accuracy on ListOps-DG (Stability test). We report the mean and standard deviation of of $10$\d. Our models were trained on lengths $\leq$ 100, depth $\leq$ 6, and arguments $\leq$ 5. Subscript represents standard deviation. As an example, $90_1 = 90\pm0.1$.}
\label{table:listopsdg_stability}
% \vspace{-4mm}
\end{table*}

\begin{table*}[t]
\small
\centering
\def\arraystretch{1.2}
\begin{tabular}{  l | l | l l l | l l |l} 
%\toprule
\hline
\textbf{Model} & \textbf{DG1} & \multicolumn{3}{c}{\textbf{Length Gen.}} & \multicolumn{2}{|c|}{\textbf{Argument Gen.}} & \textbf{LRA}\\
(Lengths) & $\leq$ 50 & 200-300 & 500-600 & 900-1000 & 100-1000 & 100-1000 & $2000$\\
(Arguments) & $\leq$ 5 & $\leq$ 5 & $\leq$ 5 & $\leq$ 5 & 10 & 15 & $\leq$ 10\\
(Depths) & 8-10 & $\leq$ 20 & $\leq$ 20 & $\leq$ 20 & $\leq$ 10 & $\leq$ 10 & $\leq$ 10\\
\hline
\multicolumn{5}{l}{\textit{After Training on ListOps-DG1}}\\
\hline
NDR (layer 24) & $96.7$ & $48.9$ & $32.85$ & $22.1$ & $65.65$ & $64.6$ & $42.6$\\
NDR (layer 48) & $91.75$ & $34.60$ & $24.05$ & $19.7$ & $54.65$ & $52.45$ & $39.95$\\
\hline
\textbf{Model} & \textbf{DG2} & \multicolumn{3}{c}{\textbf{Length Gen.}} & \multicolumn{2}{|c|}{\textbf{Argument Gen.}} & \textbf{LRA}\\
(Lengths) & $\leq$ 100 & 200-300 & 500-600 & 900-1000 & 100-1000 & 100-1000 & $2000$\\
(Arguments) & $\leq$ 5 & $\leq$ 5 & $\leq$ 5 & $\leq$ 5 & 10 & 15 & $\leq$ 10\\
(Depths) & 8-10 & $\leq$ 20 & $\leq$ 20 & $\leq$ 20 & $\leq$ 10 & $\leq$ 10 & $\leq$ 10\\
\hline
\multicolumn{5}{l}{\textit{After Training on ListOps-DG2}}\\
\hline
NDR (layer 24) & $95.6$ & $44.15$ & $30.7$ & $20.3$ & $67.85$ & $58.05$ & $46$\\
NDR (layer 48) & $92.65$ & $38.6$ & $29.15$ & $22.1$ & $73.1$ & $64.4$ & $50.4$\\
\hline
\end{tabular}
%\vspace{+.5em}
\caption{Accuracy on ListOps-DG1 and ListOps-DG2. We report the max of $3$ runs. In ListOps-DG1, NDR was trained on lengths $\leq$ 50, depth $\leq$ 6, and arguments $\leq$ 5. In ListOps-DG2, NDR was trained on lengths $\leq$ 100, depth $\leq$ 6, and arguments $\leq$ 5. Layers denote the number of layers used during inference.}
\label{table:listopsdg-NDR}
% \vspace{-4mm}
\end{table*}

\subsection{NDR Experiments}
\label{ndr_exp}
\textbf{Dataset Settings: }Neural Data Routers (NDR) is a Transformer-based model that was shown to perform well in algorithmic tasks including ListOps \citep{csordas-2022-ndr}. We tried some experiments with it too. We found NDR to be struggling in the original ListOps splits or the ListOps-DG split. We noticed that in the paper \citep{csordas-2022-ndr}, NDR was trained in a much larger sample size ($\sim 10$ times more data than in ListOps-DG) and also on lower sequence lengths ($\sim 50$). To better check for the capabilities of NDR, we created two new ListOps split - DG1 and DG2. In DG1, we set the sequence length to $10$-$50$ in training, development, and testing set. We created $1$ million data for training, and $2000$ data for development and testing. Other parameters (number of arguments, depths etc.) are the same as in ListOps-DG split. Split DG2 is the same as ListOps-DG split in terms of data-generation parameters (i.e., it includes length sizes $\leq 100$) but with much larger sample size for the training split (again, $1$ million samples same as DG1). We present the results in Table \ref{table:listopsdg-NDR}. 

\textbf{Results: }We find that even when we focus on the best of $3$ runs in the table, although NDR generalizes to slightly higher depths ($8$-$10$ from $\leq 6$) (as reported in \citep{csordas-2022-ndr}), it still struggles with splits with orders of magnitude higher depths, lengths, and unseen arguments. Following the suggestions of \cite{csordas-2022-ndr}, we also increase the number of layers during inference (e.g., up to $48$) to handle higher depth sequences but that did not help substantially. Thus, even after experiencing more data, NDR generalizes worse than Ordered Memory, CRvNN, or BT-GRC. Moreover, NDR requires some prior estimation of the true computation depth of the task for its hyperparameter setup for efficient training unlike the other latent-tree models.

\begin{table*}[t]
\small
\centering
\def\arraystretch{1.2}
\begin{tabular}{  l | l l l l l | l} 
%\toprule
\hline
\textbf{Model} & \multicolumn{5}{c}{\textbf{Number of Operations}} & \multicolumn{1}{|l}{}\\
& 8 & 9 & 10 & 11 & 12 & C\\
\hline
\multicolumn{5}{l}{\textit{With gold trees}}\\
\hline
GoldTreeGRC & $97.14_{1}$ & $96.5_{2}$ & $95.29_{2.5}$ & $94.21_{9.9}$ & $93.67_{7.7}$ & $97.41_{1.6}$\\
\hline
\multicolumn{5}{l}{\textit{Baselines without gold trees}}\\
\hline
Transformer* & $52$ & $51$ & $51$ & $51$ & $48$ & $51$\\
Universal Transformer* & $52$ & $51$ & $51$ & $51$ & $48$ & $51$\\
%RRNet* & $84$ & $81$ & $78$ & $74$ & $72$ & $71$ & ---\\
ON-LSTM* & $87$ & $85$ & $81$ & $78$ & $75$ & $60$\\
%LSTM* & $88$ & $84$ & $80$ & $78$ & $71$ & $69$ & $59$\\
Self-IRU$\dagger$ & $95$ & $93$ & $92$ & $90$ & $88$ & ---\\
RecurrentGRC & $93.04_6$ & $90.43_{4.9}$ & $88.48_6$ & $86.57_{5.8}$ & $80.58_{1.5}$ & $83.17_{5.1}$\\
GumbelTreeGRC & $93.46_{14}$ & $91.89_{19}$ & $90.33_{22}$ & $88.43_{18}$ & $85.70_{24}$ & $89.34_{29}$\\
CYK-GRC  & $96.62_{2.3}$ & $96.07_{4.6}$ & $94.67_{11}$ & $93.44_{8.8}$ & $92.54_{9.3}$ & $77.08_ {27}$\\
%BSRP & $91.74_{17}$ & $87.95_{22}$ & $85.70_{29}$ & $82.27_{36}$ & $79.67_{43}$ & $74.37_{37}$ & $77.41_{5.94}$\\
CRvNN & $96.9_{3.7}$ & $95.99_{2.8}$ & $94.51_{2.9}$ & $\mathbf{94.48_{5.6}}$ & $92.73_{15}$ & $89.79_{58}$\\
Ordered Memory & $\mathbf{97.5_{1.6}}$ & $\mathbf{96.74_{1.4}}$ & $94.95_{2}$ & $93.9_{2.2}$ & $93.36_{6.2}$ & $\mathbf{94.88_{7}}$\\
\hline
\multicolumn{5}{l}{Ours}\\
\hline
BT-GRC & $96.83_{1}$ & $95.99_{2.4}$ & $95.04_{2.3}$ & $\underline{94.29_{3.8}}$ & $93.36_{2.4}$ & $\underline{94.17_{14}}$\\
BT-GRC + OneSoft  & $\underline{97.03_{1.4}}$ & $\underline{96.49_{1.9}}$ & $\mathbf{95.43_{4.5}}$ & $94.21_{6.6}$ & $\mathbf{93.39_{1.5}}$ & $78.04_{43}$\\
\hline
\end{tabular}
%\vspace{+.5em}
\caption{Mean accuracy and standard deviaton on the Logical Inference for $\geq 8$ number of operations after training on samples with $\leq 6$ operations. We also report results of the systematicity split C. We bold the best results and underline the second-best for all models without gold trees. * indicates that the results were taken from \cite{shen-2019-ordered} and $\dagger$ indicates results from \cite{zhang-2021-self}. Our models were run $3$ times on different seeds. Subscript represents standard deviation. As an example, $90_1 = 90\pm0.1$.}
\label{table:PNLIRESULTS}
\vspace{-4mm}
\end{table*}

\subsection{Synthetic Logical Inference Results}
\label{synthlogic}
\textbf{Dataset Settings:} We also consider the synthetic testbed for detecting logical relations between sequence pairs as provided by \citet{bowman-2015-tree}. Following \citet{tran-2018-importance}, we train the models on sequences with $\leq 6$ operators and test on data with greater number of operators (here, we check for cases with $\geq 8$ operators) to check for capacity to generalize to unseen number of operators. Similar to \cite{shen-2019-ordered, chowdhury-2021-modeling}, we also train the model on the systematicity split $C$. In this split we remove any sequence matching the pattern $* ( and (not *) ) *$ from the training set and put them in the test set to check for systematic generalization. 

\textbf{Results:} In Table \ref{table:PNLIRESULTS}, in terms of the number of operations generalization, our proposed BT-Cell model performs similarly to prior SOTA models like Ordered Memory (OM) and CRvNN while approximating GoldTreeGRC for both beam sizes. In terms of systematicity (split C), OM and BT-GRC perform similarly (both above $94\%$) and much better than the other models. Surprisingly, however, OneSoft extension also hurts systematicity in this context. CYK-GRC shows promise in operator generalization but shows poor systematicity as well. GumbelTreeGRC performs better than RecurrentGRC but still far from SOTA which is not unexpected given its poor results in ListOps.

\begin{table*}[t]
\small
\centering
\def\arraystretch{1.2}
\begin{tabular}{  l | l | l l l | l l |l} 
%\toprule
\hline
\textbf{Model} & \textbf{near-IID} & \multicolumn{3}{c}{\textbf{Length Gen.}} & \multicolumn{2}{|c|}{\textbf{Argument Gen.}} & \textbf{LRA}\\
(Lengths) & $\leq$ 1000 & 200-300 & 500-600 & 900-1000 & 100-1000 & 100-1000 & $2000$\\
(Arguments) & $\leq$ 5 & $\leq$ 5 & $\leq$ 5 & $\leq$ 5 & 10 & 15 & 10\\
\hline
RandomTreeGRC & $70.56$ & $48.70$ & $45.35$ & $37.53$ &  $54.8$ & $55.6$ & $49.8$\\
BalancedTreeGRC & $59.4$ & $44.85$ & $43.35$ & $35.70$ & $45.88$ & $45.25$ & $41.95$\\
\hline
\end{tabular}
\caption{Accuracy of different models on ListOps (same setting as in Table \ref{table:listops}). We report the median of $3$ runs.}
\label{table:listops_heuristics}
\end{table*}

\begin{table*}[t]
\small
\centering
\def\arraystretch{1.2}
\begin{tabular}{  l | l | l l | l l l l l l} 
%\toprule
\hline
\textbf{Model} & \textbf{SST5} & \multicolumn{2}{|c|}{\textbf{IMDB}} & \multicolumn{6}{|c}{\textbf{MNLI}} \\
& IID & Con. & Count. & \textbf{M} & \textbf{MM} & \textbf{Len M} & \textbf{Len MM} & \textbf{Neg M} & \textbf{Neg MM} \\
\hline
\multicolumn{10}{c}{Heuristic Trees}\\
\hline
RandomTreeGRC & $51.78_{1.2}$ & $74.93_{14}$ & $82.38_{9.3}$ & $72.2_3$ & $72.3_5$ & $61.4_{23}$ & $62.3_{23}$ & $51.7_3$ & $52.7_7$\\
BalancedTreeGRC & $52.35_{6.2}$ & $74.93_{22}$ & $83.61_{15}$\ & $71.1_5$ & $71.4_1$ & $59_8$ & $60.7_5$ & $50.2_4$ & $50.4_6$\\
\hline
\multicolumn{10}{c}{Beam Size $2$}\\
\hline
BT-GRC & $52.14_{2.8}$ & $75.21_{23}$ & $82.51_{23}$ & $\mathbf{72.6_1}$ & $\mathbf{72.6_2}$ & $\mathbf{66.6_5}$ & $\mathbf{68.1_6}$ & $\mathbf{53.3_{21}}$ & $\mathbf{54.4_{24}}$\\
BT-GRC + OneSoft & $\mathbf{52.2_{4.4}}$ & $\mathbf{75.89_{22}}$ & $\mathbf{85.45_{15}}$ & $71.1_3$ & $71.9_1$ & $63.7_{17}$ & $65.6_{12}$ & $51.8_{19}$ & $53_{12}$\\
\hline
\end{tabular}
%\vspace{+.5em}
\caption{Mean accuracy and standard deviaton on SST5, IMDB, and MNLI. Con. represents Contrast set and Count. represents Countefactuals. Our models were run $3$ times on different seeds. Subscript represents standard deviation. As an example, $90_1 = 90\pm0.1$.}
\label{table:real_heuristics}
\vspace{-4mm}
\end{table*}

\begin{table*}[t]
\small
\centering
\def\arraystretch{1.2}
\begin{tabular}{  l | l | l l l | l l |l} 
%\toprule
\hline
\textbf{Model} & \textbf{near-IID} & \multicolumn{3}{c}{\textbf{Length Gen.}} & \multicolumn{2}{|c|}{\textbf{Argument Gen.}} & \textbf{LRA}\\
(Lengths) & $\leq$ 1000 & 200-300 & 500-600 & 900-1000 & 100-1000 & 100-1000 & $2000$\\
(Arguments) & $\leq$ 5 & $\leq$ 5 & $\leq$ 5 & $\leq$ 5 & 10 & 15 & 10\\
\hline
\multicolumn{5}{l}{\textit{With gold trees}}\\
\hline
GoldTreeGRC & $99.9_{.2}$ & $99.9_{.9}$ & $99.8_1$ & $100_{.5}$ & $81.2_{28}$ & $79.5_{14}$ & $78.5_{29}$\\
\hline
\multicolumn{5}{l}{\textit{Baselines without gold trees}}\\
\hline
RecurrentGRC & $83.8_{4.5}$ & $33.9_{1.2}$ & $18.1_{38}$ & $13.7_{25}$ & $37.6_{18}$ & $29.2_{30}$ & $18.5_{35}$\\
GumbelTreeGRC & $75_{4.6}$ & $47.7_{8.4}$ & $42.7_{2.8}$ & $37.4_{42}$ & $50.9_{15}$ & $51.4_{16}$ & $45.3_{12}$\\
Ordered Memory & $\mathbf{99.9_{.3}}$ & $\mathbf{99.6_{.7}}$ & $92.4_{13}$ & $76.3_{13}$ & $\mathbf{83.2_{24}}$ & $\mathbf{76.3_{38}}$ & $\mathbf{79.3_{18}}$ \\
%CRvNN$\dagger$ & $99.6_3$ & $98.51_{11}$ & $97.95_{11}$ & $96.78_{19}$ & --- & --- & ---\\
%CRvNN & $98.86$ & $95.89$ & $93.15$ & $89.4$ & $57.8$ & $24.35$ & $45.1$ \\
CRvNN &  $\underline{99.7_{2.8}}$ & $98.8_{11}$ & $\mathbf{97.2_{23}}$ & $\mathbf{94.9_{49}}$ & $66.6_{40}$ & $43.7_{38}$ & $55.38_{44}$\\
\hline
\multicolumn{5}{l}{\textit{Ours}}\\
\hline
BT-GRC & $99.4_{2.7}$ & $\underline{96.8_{10}}$ & $93.6_{22}$ & $88.4_{27}$ & $\underline{75.2_{28}}$ & $59.1_{79}$ & $63.4_{57}$\\
BT-GRC + OneSoft & $99.6_{5.4}$ & $97.2_{35}$ & $\underline{94.8_{65}}$ & $\underline{92.2_{86}}$ & $73.3_{64}$ & $\underline{63.1_{92}}$ & $\underline{66.1_{101}}$\\
\hline
\end{tabular}
\caption{Accuracy on ListOps. For our models we report the mean of $3$ runs. Our models were trained on lengths $\leq$ 100, depth $\leq$ 20, and arguments $\leq$ 5. We bold the best results and underline the second-best among models that do not use gold trees. Subscript represents standard deviation. As an example, $90_1 = 90\pm0.1$.}
\label{table:listops-mean}
\vspace{-4mm}
\end{table*}

\subsection{Beam 2 results on Natural Language Tasks}
\label{real2}
In Table \ref{table:real_heuristics}, we report some additional results (particularly on natural language tasks) with beam width $2$. The results here are a bit mixed and OneSoft does not consistently demonstrate superiority over base BT-GRC but generally they are close to each other.

\subsection{Heuristics Tree Models}
\label{heuristics}
We consider two heuristics-based tree models - RandomTreeGRC (uses random tree strcutures) and BalancedTreeCell (follows a balanaced binary tree structure \cite{shi-etal-2018-tree}). We run them on ListOps and show the results in Table \ref{table:listops_heuristics}. As we would expect the heuristics-based model perform very poorly. We also run them on natural language data and show the results in Table \ref{table:real_heuristics}. It performs relatively more competitively against other models in realistic data but they still generally fall behind the OM, CRvNN, and BT-Cell.

\subsection{Mean Results on ListOps}
\label{mean}
In Table \ref{table:listops-mean}, we report the mean results of our model on the original ListOps splits. Although CRvNN outperforms BT-GRC + OneSoft slightly on the mean length generalization performance, it is still $10-20\%$ behind BT-GRC + OneSoft in argument generalization. 

\begin{table*}[t]
\small
\centering
\def\arraystretch{1.2}
\begin{tabular}{  l | l } 
%\toprule
\hline
\textbf{Score} & \textbf{Parsed Structures}\\
\hline
\multicolumn{2}{c}{\textbf{BT-GRC (beam size 5)}}\\ 
\hline

$0.42$ & ((i (did not)) (((like a) (single minute)) ((of this) film)))\\
$0.40$ & (((i (did not)) ((like a) (single minute))) ((of this) film))\\
$0.20$ & ((i (did not)) (((like a) ((single minute) of)) (this film)))\\
\hline
$0.40$ &  ((i (shot an)) ((elephant in) (my pajamas)))\\
$0.21$ & (((i shot) (an elephant)) ((in my) pajamas))\\
$0.19$ & (((i shot) (an elephant)) (in (my pajamas)))\\
$0.19$ & ((i shot) ((an elephant) ((in my) pajamas)))\\
\hline
$0.40$ & ((john saw) ((a man) (with binoculars)))\\
$0.40$ & (((john saw) (a man)) (with binoculars))\\
$0.20$ &  ((john (saw a)) ((man with) binoculars))\\
\hline
$0.61$ & (((roger (dodger is)) (one (of the))) (((most compelling) (variations of)) (this theme)))\\
$0.40$ & (((roger (dodger is)) ((one (of the)) (most compelling))) ((variations of) (this theme)))\\ 
\hline
\multicolumn{2}{c}{\textbf{BT-GRC (beam size 2)}}\\ 
\hline
$0.50$ & ((i ((did not) like)) (((a single) minute) ((of this) film)))\\
$0.50$ &  ((i (((did not) like) (a single))) ((minute of) (this film)))\\
\hline
$0.50$ & ((i (shot an)) ((elephant in) (my pajamas)))\\
$0.50$ & ((i ((shot an) elephant)) ((in my) pajamas))\\
\hline
$0,51$ & ((john (saw a)) ((man with) binoculars))\\
$0.49$ & (john (((saw a) man) (with binoculars)))\\
\hline
$1.0$ & ((roger ((dodger is) one)) ((((of the) most) (compelling variations)) ((of this) theme)))\\
\hline

\end{tabular}
%\vspace{+.5em}
\caption{Parsed Structures of BT-GRC trained on MNLI. Each block represents different beams.}.
\label{table:btgrc_mnli_parse}
\vspace{-4mm}
\end{table*}

\begin{table*}[t]
\small
\centering
\def\arraystretch{1.2}
\begin{tabular}{  l | l } 
%\toprule
\hline
\textbf{Score} & \textbf{Parsed Structures}\\
\hline
\multicolumn{2}{c}{\textbf{BT-GRC + OneSoft (beam size 5)}}\\ 
\hline

$0.42$ & (((i did) (not like)) (((a single) minute) ((of this) film)))\\
$0.20$ &  ((((i did) not) ((like a) single)) ((minute of) (this film)))\\
$0.19$ & ((((i did) (not like)) ((a single) minute)) ((of this) film))\\
$0.19$ & (((i (did not)) ((like a) single)) ((minute of) (this film)))\\
\hline
$0.41$ & (((i shot) an) ((elephant in) (my pajamas)))\\
$0.21$ & (((i shot) (an elephant)) ((in my) pajamas))\\
$0.19$ & ((i (shot an)) ((elephant in) (my pajamas)))\\
$0.19$ & ((((i shot) an) (elephant in)) (my pajamas))\\
\hline
$0.21$ & ((john (saw a)) ((man with) binoculars))\\
$0.20$ &  (((john saw) (a man)) (with binoculars))\\
$0.20$ & ((john saw) ((a man) (with binoculars)))\\
$0.19$ & ((john ((saw a) man)) (with binoculars))\\
\hline
$0.40$ & (((roger dodger) (is one)) ((((of the) most) (compelling variations)) ((of this) theme)))\\
$0.21$ &  (((roger (dodger is)) ((one of) the)) (((most compelling) variations) ((of this) theme)))\\
$0.20$ & ((((roger dodger) (is one)) ((of the) most)) ((compelling variations) ((of this) theme)))\\
$0.19$ & ((roger (dodger is)) ((((one of) the) ((most compelling) variations)) ((of this) theme)))\\
\hline
\multicolumn{2}{c}{\textbf{BT-GRC + OneSoft (beam size 2)}}\\ 
\hline
$0.57$ & ((i ((did not) like)) (((a single) minute) ((of this) film)))\\
$0.43$ & ((i ((did not) like)) (((a single) (minute of)) (this film)))\\
\hline
$0.54$ & ((i ((shot an) elephant)) ((in my) pajamas))\\
$0.46$ &  ((i (shot an)) ((elephant in) (my pajamas)))\\
\hline
$0.55$ & ((john (saw a)) ((man with) binoculars))\\
$0.45$ & ((john ((saw a) man)) (with binoculars))\\
\hline
$0.53$ &  ((roger ((dodger is) one)) ((((of the) most) (compelling variations)) ((of this) theme)))\\
$0.47$ & (((roger ((dodger is) one)) ((of the) most)) ((compelling variations) ((of this) theme)))\\
\hline

\end{tabular}
%\vspace{+.5em}
\caption{Parsed Structures of BT-GRC + OneSoft trained on MNLI. Each block represents different beams.}.
\label{table:btgrc_softpath_mnli_parse}
\vspace{-4mm}
\end{table*}

\begin{table*}[t]
\small
\centering
\def\arraystretch{1.2}
\begin{tabular}{  l | l } 
%\toprule
\hline
\textbf{Score} & \textbf{Parsed Structures}\\
\hline
\multicolumn{2}{c}{\textbf{BT-GRC (beam size 5)}}\\ 
\hline
$0.26$ & ((((((i (((did not) like) a)) single) minute) of) this) film)\\
$0.23$ & ((((i ((did (((not like) a) single)) minute)) of) this) film)\\
$0.19$ & (i (did ((not ((like (((a single) minute) of)) this)) film)))\\
$0.18$ & (((((((((i did) not) like) a) single) minute) of) this) film)\\
$0.14$ & (((((i (((did not) like) a)) single) minute) of) (this film))\\
\hline
$0.24$ & (i (shot (an (elephant ((in my) pajamas)))))\\
$0.22$ & ((((((i shot) an) elephant) in) my) pajamas)\\
$0.21$ & (i (shot (an (((elephant in) my) pajamas))))\\
$0.19$ & (((i (((shot an) elephant) in)) my) pajamas)\\
$0.14$ & (i ((shot (((an elephant) in) my)) pajamas))\\
\hline
$0.25$ & (john (saw (a (man (with binoculars)))))\\
$0.22$ & (((((john saw) a) man) with) binoculars)\\
$0.21$ & (john (saw (a ((man with) binoculars))))\\
$0.18$ & (john (saw (((a man) with) binoculars)))\\
$0.15$ & ((john (((saw a) man) with)) binoculars)\\
\bottomrule
\end{tabular}
%\vspace{+.5em}
\caption{Parsed Structures of BT-GRC trained on SST5. Each block represents different beams.}.
\label{table:btgrc_sst5_parse}
\vspace{-4mm}
\end{table*}

\subsection{Parse Tree Analysis}
\label{parse_analysis}
In this section, we analyze the induced structures of BT-Cell models. Note, however, although induced structures can provide some insights into the model, we can draw limited conclusions from them. First, if we take a stance similar to \cite{choi-2018-learning} in considering it suitable to allow different kinds of structures to be induced as appropriate for a specific task then it is not clear how structures should be evaluated by themselves (besides just the donwstream task evaluations). Second, the extracted structures may not completely reflect what the models may implicitly induce because the recursive cell can override some of the parser decisions (given how there is evidence that even simple RNNs \cite{bowman-2015-tree} can implicitly model different tree structures within its hidden states to an extent even when its explicit structure always conform to the left-to-right order of composition). Third, even if the extracted structure perfectly reflects what the model induces, another side of the story is the recursive cell itself and how it utilizes the structure for language understanding. This part of the story can still remain unclear because of the blackbox-nature of neural nets. Nevertheless, extractive structures may still provide some rough idea of the inner workings of the BT-Cell variants. 

In Table \ref{table:btgrc_mnli_parse}, we show the parsed structures of some iconic sentences by BT-GRC after it is trained on MNLI. We report all beams and their corresponding scores. Note, although beam search ensures that the sequence of parsing actions for each beam is unique, different sequences of parsing action can still lead to the same structure. Thus, some beams end up being duplicates. In such cases, for the sake of more concise presentation, we collapse the duplicates into a single beam and add up their corresponding scores. This is why we can note in Table \ref{table:btgrc_mnli_parse} that we sometimes have fewer induced structures than the beam size.  

At a rough glance, we can see that the different induced structures roughly correspond to human intuitions. One interesting appeal for beam search is that it can more explicitly account for ambiguous interpretations corresponding to ambiguous structures. For example, \textit{``i shot an elephant in my pajamas"} is ambiguous with respect to whether it is the elephant who is in the shooter's pajamas, or if it is the shooter who is in the pajamas. The induced structure (beam size $5$ model in Table \ref{table:btgrc_mnli_parse}) \textit{(((i shot) (an elephant)) ((in my) pajamas))} corresponds better to the latter interpretation whereas \textit{((i shot) ((an elephant) ((in my) pajamas)))} corresponds better to the former interpretation (because ``an elephant" is first composed with ``in my pajamas"). 

Similar to above, \textit{``john saw a man with binoculars"} is also ambiguous. Its interpretation is ambiguous with respect to whether it is John who is seeing through binoculars, or whether it is the man who just possesses the binoculars. Here, again, we can find (beam size $5$ model in Table \ref{table:btgrc_mnli_parse}) that the induced structure \textit{(((john saw) (a man)) (with binoculars)} corresponds better to the former interpretation whereas \textit{((john saw) ((a man) (with binoculars)))} corresponds better to the latter. Generally, we find the score distributions to have a high entropy. A future consideration would be whether we should add an auxiliary objective to minimize entropy. 

In Table \ref{table:btgrc_softpath_mnli_parse}, we show the parsed structures of the same sentences by BT-GRC + OneSoft after it is trained on MNLI. Most of the points above applies here for OneSoft as well. Interestingly, OneSoft seems to have a relatively lower entropy distribution - that is most evident in beam size $2$. 

We found the structures induced by BT-Cell variants after training on SST5 or IMDB to be more ill-formed. This may indicate that sentiment classification does not provide a strong enough signal for parsing or rather, exact induction of structures are not as necessary \citep{iyyer-etal-2015-deep}. We show the parsings of these models after training on IMDB and SST datasets on GitHub\footnote{\url{https://github.com/JRC1995/BeamTreeRecursiveCells/blob/main/parses.txt}}. A few examples are presented also in Table \ref{table:btgrc_sst5_parse}.  

\section{Extended Related Work}
\label{ext_related_works}
Initially RvNN \citep{pollack-1990-recursive,goller1996learning, Socher10learningcontinuous} was used with user-annotated tree-structured data. Some works explored the use of heuristic trees such as balanced trees for RvNN-like settings \citep{munkhdalai-yu-2017-neural, shi-etal-2018-tree}. \citet{le-zuidema-2015-compositional,tai-etal-2015-improved,zhu2015lstms,zhu-etal-2016-dag} explored the incorporation of Long Short Term Memory Cells \cite{hochreiter1997long} with RvNNs. In due time, several approaches were introduced for dynamically inducing structures from data for RvNN-style processing. This includes the greedy easy-first framework using children-reconstruction loss \citep{socher-etal-2011-semi}, gumbel softmax \citep{choi-2018-learning}, or SPIGOT \cite{peng-etal-2018-backpropagating}, Reinforcement Learning-based frameworks \citep{havrylov-2019-cooperative}, CYK-based framework \citep{le-zuidema-2015-forest, maillard_clark_yogatama_2019, drozdov-etal-2019-unsupervised-latent, hu-etal-2021-r2d2}, shift-reduce parsing or memory-augmented or stack-augmented RNN frameworks \citep{Grefenstette2015learning, bowman-etal-2016-fast, yogatama2017learning, maillard-clark-2018-latent, shen-2019-ordered, dusell-chiang-2020-learning, dusell2022learning}, and soft-recursion-based frameworks \citep{chowdhury-2021-modeling, zhang-2021-self}. Besides RvNNs, other approaches range from adding information-ordering biases to hidden states in RNNs \citep{shen-2018-ordered} or even adding additional structural or recursive constraints to Transformers \citep{wang2019tree,Nguyen2020Tree-Structured, fei2020retrofitting,shen-2021-structformer, csordas-2022-ndr}. 

%There are multiple versions of differentiable top-k operators or sorting functions \citep{adams-2011-ranking,grover2018stochastic,cuturi-2019-differentiable,xie-2020-differentiable,blondel2020fast, peterson-2021-differentiable, petersen2022monotonic}. We leave a more exhaustive analysis of them as future work. However, note that many of them would suffer from the same systematic issues as SOFT top-k \cite{xie-2020-differentiable} - that is they can significantly slow down the model and they can lead to ``washed out" representations.

\section{Hyperparameters}
\label{hyperparameters}
For all recursive/recurrent  models, we use a linear layer followed by layer normalization for initial leaf transformation before starting the recursive loop (similar to \citet{shen-2019-ordered, chowdhury-2021-modeling}). Overall we use the same boilerplate classifier architecture for classification and the same boilerplate sentnece-pair siamese architecture for logical inference as \cite{chowdhury-2021-modeling} over our different encoders.  In practice, for BT-Cell, we use a stochastic top-$k$ through gumbel perturbation similar to \citet{kool2019stochastic}. However, we find deterministic selection to work similarly. In our implementation of CRvNN, we ignore some extraneous elements from CRvNN such as transition features and halt penalty which were deemed to have little effect during ablation in \citet{chowdhury-2021-modeling}.

In terms of the optimizer, hidden size, and other hyperparameters besides dropout, we use the same ones as used by \citep{chowdhury-2021-modeling} for all models for corresponding datasets; for number of memory slots and other ordered memory specific parameters we use the same ones as used by \citep{shen-2019-ordered}. Generally, we use a patience of $5$ for the original ListOps training for all models, but we use a patience of $10$ for CRvNN (same as used in \citet{chowdhury-2021-modeling}) to replicate a length generalization performance closer to that reported in \citet{chowdhury-2021-modeling}. For BSRP-Cell we use a beam size of $8$ (we also tried with $5$ but results were similar or slightly worse). We use a dropout rate of $0.1$ for logical inference for all models (tuned on the validation set using grid search among $[0.1, 0.2, 0.3, 0.4]$ with $5$ epochs per run using BalancedTreeCell for GRC-based models and GumbelTreeLSTM for LSTM based models). We use dropouts in the same places as used in \citep{chowdhury-2021-modeling}.  We then use the same chosen dropouts for ListOps. We tune the dropouts for SST in the same way (but with a maximum epoch of $20$ per trial) on SST5 using RecurrentGRC for GRC-models, and Gumbel-Tree-LSTM for LSTM models. After tuning, for GRC-based models in SST5, we found a dropout rate of $0.4$ for input/output dropout layers, and $0.2$ for the dropout layer in the cell function. We found a dropout of $0.3$ for LSTM-based models in SST5. and  We share the hyperparameters of SST5 with IMDB. For MNLI, we used similar settings as \citet{chowdhury-2021-modeling}. 

For NDR experiments, we use the same hyperparameters as used for ListOps by \citet{csordas-2022-ndr}. The hyperparameters will also be available with the code. 
All experiments are run in a single RTX A6000 GPU. 
%%%%%%%%%%%%%%%%%%%%%%%%%%%%%%%%%%%%%%%%%%%%%%%%%%%%%%%%%%%%%%%%%%%%%%%%%%%%%%%
%%%%%%%%%%%%%%%%%%%%%%%%%%%%%%%%%%%%%%%%%%%%%%%%%%%%%%%%%%%%%%%%%%%%%%%%%%%%%%%

\end{document}